\documentclass[final,authoryear,5p,times,twocolumn]{elsarticle}
\pdfoutput=1

\usepackage{graphicx}
\usepackage{subfig}
\usepackage[cmex10]{amsmath}
\usepackage{amssymb}
\usepackage{multirow}
\usepackage{url}
\usepackage{placeins}
\usepackage{floatrow}
\usepackage{color}

\newcommand{\argmax}{\mathop{\mathrm{argmax}}}

\journal{Medical Image Analysis}

\begin{document}

\begin{frontmatter}



\title{Deep Image Mining for Diabetic Retinopathy Screening}

\author[label1]{Gwenol\'e~Quellec\corref{cor1}}
\ead{gwenole.quellec@inserm.fr}
\cortext[cor1]{LaTIM - B\^atiment 1 - CHRU Morvan - 2, Av. Foch\\29609 Brest CEDEX - FRANCE\\Tel.: +33 2 98 01 81 29 / Fax: +33 2 98 01 81 24}
\author[label2,label1]{Katia~Charri\`ere}
\author[label2,label1]{Yassine~Boudi}
\author[label3,label1,label4]{B\'eatrice~Cochener}
\author[label3,label1]{Mathieu~Lamard}
\address[label1]{Inserm, UMR 1101, Brest, F-29200 France}
\address[label2]{Institut Mines-Telecom; Telecom Bretagne, Dpt ITI, Brest, F-29200 France}
\address[label3]{Univ Bretagne Occidentale, Brest, F-29200 France}
\address[label4]{Service d'Ophtalmologie, CHRU Brest, Brest, F-29200 France}

\begin{abstract}
Deep learning is quickly becoming the leading methodology for medical image analysis. Given a large medical archive, where each image is associated with a diagnosis, efficient pathology detectors or classifiers can be trained with virtually no expert knowledge about the target pathologies. However, deep learning algorithms, including the popular ConvNets, are black boxes: little is known about the local patterns analyzed by ConvNets to make a decision at the image level. A solution is proposed in this paper to create heatmaps showing which pixels in images play a role in the image-level predictions. In other words, a ConvNet trained for image-level classification can be used to detect lesions as well. A generalization of the backpropagation method is proposed in order to train ConvNets that produce high-quality heatmaps. The proposed solution is applied to diabetic retinopathy (DR) screening in a dataset of almost 90,000 fundus photographs from the 2015 Kaggle Diabetic Retinopathy competition and a private dataset of almost 110,000 photographs (e-ophtha). For the task of detecting referable DR, very good detection performance was achieved: $A_z = 0.954$ in Kaggle's dataset and $A_z = 0.949$ in e-ophtha. Performance was also evaluated at the image level and at the lesion level in the DiaretDB1 dataset, where four types of lesions are manually segmented: microaneurysms, hemorrhages, exudates and cotton-wool spots. For the task of detecting images containing these four lesion types, the proposed detector, which was trained to detect referable DR, outperforms recent algorithms trained to detect those lesions specifically, with pixel-level supervision. At the lesion level, the proposed detector outperforms heatmap generation algorithms for ConvNets. This detector is part of the Messidor\textsuperscript{\textregistered} system for mobile eye pathology screening. Because it does not rely on expert knowledge or manual segmentation for detecting relevant patterns, the proposed solution is a promising image mining tool, which has the potential to discover new biomarkers in images.
\end{abstract}


\end{frontmatter}

\section{Introduction}
\label{sec:Introduction}

\begin{figure*}
  \begin{center}
      \includegraphics[width=0.6\textwidth]{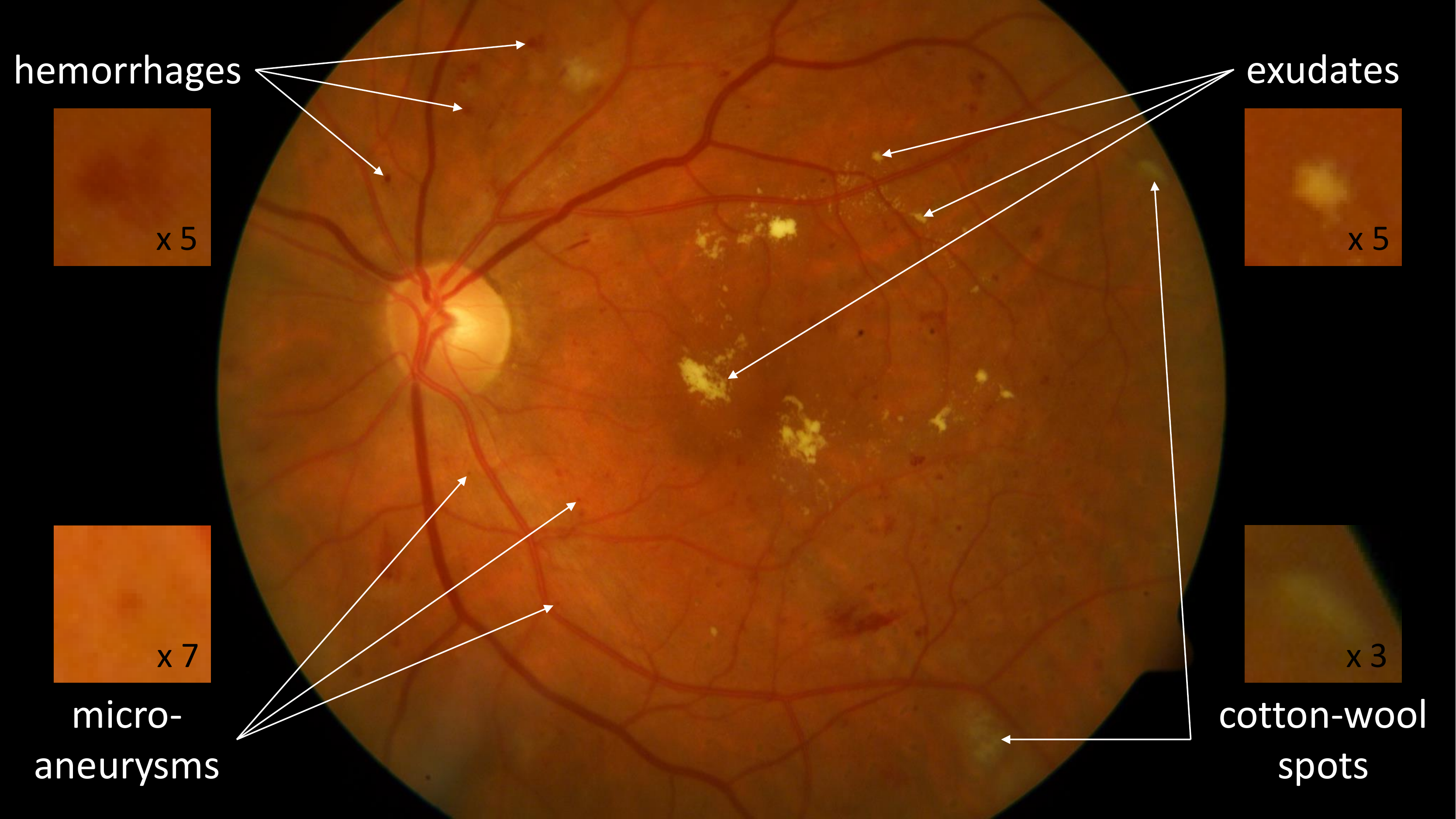}
  \end{center}
  \caption{Most common lesions caused by diabetic retinopathy (image015 from the DiaretDB1 dataset with magnified lesions).}
  \label{fig:lesionsDR}
\end{figure*}

Retinal pathologies are responsible for millions of blindness cases worldwide. The leading causes of blindness are glaucoma (4.5 million cases), age-related macular degeneration (3.5 million cases) and diabetic retinopathy (2 million cases).\footnote{\url{www.who.int/blindness/causes/priority}} Early diagnosis is the key to slowing down the progression of these diseases and therefore preventing the occurrence of blindness. In the case of diabetic retinopathy (DR) screening, diabetic patients have their retinas examined regularly: a trained reader searches for the early signs of the pathology in fundus photographs (see Fig. \ref{fig:lesionsDR}) and decides whether the patient should be referred to an ophthalmologist for treatment. In order to reduce the workload of human interpretation, and therefore streamline retinal pathology screening, various image analysis algorithms have been developed over the last few decades. The first solutions were trained to detect lesions (at the pixel level) using manual segmentations (at the pixel level) for supervision \citep{winder_algorithms_2009, abramoff_retinal_2010}: this is what we call computer-aided detection (CADe) algorithms. Based on the detected lesions, other algorithms were trained to detect pathologies (at the image level) \citep{abramoff_retinal_2010}: this is what we call computer-aided diagnosis (CADx) algorithms. In recent years, new algorithms were designed to detect pathologies directly, using diagnoses (at the image level) only for supervision: these algorithms are based on multiple-instance learning \citep{quellec_multiple-instance_2017, manivannan_sub-category_2017} or deep learning \citep{lecun_deep_2015}. Because manual segmentations are not needed, such algorithms can be trained with much larger datasets, such as anonymized archives of examination records. The next challenge is to detect lesions using diagnoses only for supervision. Besides access to large training datasets, such an approach would allow discovery of new biomarkers in images, since algorithms are not limited by the subjectivity of manual segmentations. A few multiple-instance learning algorithms, supervised at the image level, can already detect lesions \citep{melendez_novel_2015, quellec_multiple-instance_2016}. However, to our knowledge, no deep learning algorithm was designed to solve this task. The primary objective of this study is to find a way to detect lesions, or other biomarkers of DR, using deep learning algorithms supervised at the image level. A secondary objective is to use these local detections to improve DR detection at the image level.

In 2015, a machine learning competition was organized with the goal to design an automated system for grading the severity of diabetic retinopathy (DR) in images.\footnote{\url{https://www.kaggle.com/c/diabetic-retinopathy-detection}} Images were trained and evaluated at the image level in a publicly-available dataset of almost 90,000 images provided by EyePACS \citep{cuadros_eyepacs:_2009}, a free platform for DR screening. The performance criterion was the inter-rater agreement between the automated predictions and the predictions of human readers. As usual in recent pattern recognition competitions \citep{russakovsky_imagenet_2015}, the top-ranking solutions all relied on deep learning. More precisely, they relied on ensembles of ConvNets. ConvNets are artificial neural networks where each neuron only processes one portion of the input image \citep{lecun_deep_2015}. The main building-block of ConvNets are convolutional layers. In those layers, input images are convolved with multiple filters inside a sliding window (tens or hundreds of 3 $\times$ 3 to 5 $\times$ 5 multichannel filters, typically). After nonlinear post-processing, one activation map is obtained per filter. Those activation maps can be further processed by another convolutional layer, or can be nonlinearly down-sampled by a pooling layer. After several convolutional and pooling layers (10 to 30 layers, typically), ConvNets usually end with dense layers, which produce image-level predictions. Overall, ConvNets usually have a few million free parameters. The inter-rater agreement achieved by solutions of the challenge was clearly at the level of the inter-rater agreement among human readers \citep{barriga_assessing_2014}. However, many clinicians would not trust a black box, like a ConvNet (not to mention an ensemble of ConvNets), if their patient's health and their liability are at stake. Therefore, we are looking for a solution that jointly detects referable DR at the image level and detects biomarkers of this disease at the pixel level. Because of their good performance, solutions of the Kaggle DR challenge were reused and modified to also provide pixel-level detections. The proposed modifications rely on ConvNet visualization techniques. The resulting solution is part of the Messidor\textsuperscript{\textregistered} system for eye pathology screening,\footnote{\url{http://www.retinoptic.fr/}} which integrates a mobile non-mydriatic retinograph and algorithms for automated or computer-aided diagnosis.

The remaining of this paper is organized as follows. Section \ref{sec:StateArt} reviews the state of the art from an application point of view, namely deep learning for retinal image analysis, and from a methodological point of view, namely visualizing what ConvNets are learning. Section \ref{sec:HeatmapGeneration} describes the proposed lesion detection criterion. This criterion is improved by a novel optimization process in section \ref{sec:HeatmapOptimization}. Section \ref{sec:Experiments} presents experiments in three retinal image datasets (Kaggle, DiaretDB1 and e-ophtha). We end with a discussion and conclusions in section \ref{sec:DiscussionConclusions}.

\section{State of the Art}
\label{sec:StateArt}

\subsection{Deep Learning for Retinal Image Analysis}

Deep learning was recently applied to various tasks related to retinal image analysis. For landmark segmentation and lesion detection, it was applied at a pixel level. For pathology detection, it was applied at an image level.

At a pixel level, a few algorithms were proposed for segmenting retinal vessels \citep{maji_deep_2015, maji_ensemble_2016, li_cross-modality_2016} and the optic disc \citep{lim_integrated_2015, srivastava_using_2015}; others were proposed for detecting microaneurysms \citep{haloi_improved_2015}, hemorrhages \citep{van_grinsven_fast_2016} and exudates \citep{prentasic_detection_2015}, three lesions related to DR; another one was proposed for detecting various lesions (including hemorrhages and exudates) and normal anatomical structures \citep{abramoff_improved_2016}. First, \citet{maji_deep_2015, maji_ensemble_2016} use an ensemble of ConvNets to classify each pixel as `part of a vessel' or `not part of a vessel'. Similarly, \citet{lim_integrated_2015} use a ConvNet to classify each pixel as `part of the optic cup', `part of the optic disc minus the optic cup' or `not part of the optic disc'. Alternatively, \citet{srivastava_using_2015} use a network composed of (unsupervised) stacked auto-encoders followed by a supervised layer to classify each pixel as `part of the optic disc' or `not part of the optic disc'. For lesion detection, \citet{haloi_improved_2015}, \citet{van_grinsven_fast_2016}, and \citet{prentasic_detection_2015} use a ConvNet to classify pixels as `part of a target lesion (a microaneurysm, a hemorrhage, or an exudate, respectively)' or `not part of a target lesion'. \citet{abramoff_improved_2016} also use ConvNets to detect lesions or normal anatomical structures. In those seven algorithms, each pixel is classified through the analysis of a squared region centered on the pixel. In contrast, full images are analyzed in \citet{li_cross-modality_2016}: stacked auto-encoders trained on `fundus photograph' / `vessel segmentation map' pairs are used to generate vessel probability maps the size of fundus photographs. All those algorithms require manual segmentations of training images for supervision.

At an image level, algorithms were proposed for detecting glaucoma \citep{chen_glaucoma_2015, chen_automatic_2015},  age-related macular degeneration (AMD) \citep{burlina_detection_2016} and retinopathy of prematurity (ROP) \citep{worrall_automated_2016}. While \citet{lim_integrated_2015} detect glaucomatous patients using the standard cup-to-disc ratio, derived from their deep-learning-based segmentations of the optic disc and cup, \citet{chen_glaucoma_2015, chen_automatic_2015} directly classify an image as `glaucomatous' or `non-glaucomatous' through the analysis of a large region of interest centered on the optic disc, using one or two ConvNets. To detect AMD, \citet{burlina_detection_2016} use the OverFeat features, derived from a ConvNet trained on the very large, general-purpose ImageNet dataset: these features are used to train a linear support-vector machine (SVM). To detect ROP, \citet{worrall_automated_2016} fine-tuned the GoogLeNet network, also pre-trained on ImageNet. Finally, \citet{arunkumar_multi-retinal_2015} proposed an algorithm for differentiating multiple pathologies: AMD, DR, macula pucker, retinoblastoma, retinal detachment and retinitis pigmentosa. Similarly to the AMD detector, image features are extracted with a ConvNet and a multi-class SVM is used to differentiate the various pathologies.

Finally, \citet{colas_deep_2016} and \citet{gulshan_development_2016} also use deep learning techniques for detecting referable DR. In \citet{colas_deep_2016}, lesions are first detected (at the pixel level) and then DR severity is graded (at the image level). In \citet{gulshan_development_2016}, the presence of referable DR is detected at the image level using an ensemble of ConvNets.

\subsection{Visualizing what ConvNets are Learning}

Because ConvNets are black boxes, many solutions have been proposed to visualize what they have learned. The earliest solutions consisted in visualizing the trained filters or intermediate activation maps. Given the large number of convolutional units in a ConvNet, it is hard to find out from a visual inspection which pattern each of them is looking for. One way to address this issue is to find which image, inside a large dataset, maximally activates each convolutional unit \citep{girshick_rich_2014} or to generate an artificial image that maximally activates it \citep{yosinski_understanding_2015}. Besides understanding the role of each convolutional unit, an additional question arises when training ConvNet at the image level: which regions or pixels, inside the image, play a role in the image-level prediction? A simple solution was first proposed by \citet{zeiler_visualizing_2014}: portions of the image are successively masked out with a sliding window and the image-level responses are analyzed: if a relevant image area is masked out, image-level recognition performance should decrease. This approach has several limitations: 1) from a computational point of view, images need to be processed many times and, more importantly, 2) redundant objects will not be detected. Typically, if a medical image contains several similar lesions, masking a single lesion out may not affect the diagnosis.

In our field, \citet{worrall_automated_2016} proposed a visualization technique for ROP detection. This technique takes advantage of one specificity of GoogLeNet: activation values from all spatial locations in the deepest activation maps are averaged and the resulting average vector (one value per map) is processed with a softmax classifier. By removing the global average operator and applying the softmax classifier to each spatial location, relevant regions can be highlighted. The GoogLeNet network was modified to increase the definition of this visualization map from $7 \times 7$ pixels to $31 \times 31$ pixels, but this solution cannot provide pixel-level information, which can be a limitation when small lesions are involved.

In contrast, a set of methods was proposed to quantify how much each pixel impacts the image-level prediction, while analyzing the full image: the deconvolution method \citep{zeiler_visualizing_2014}, the sensitivity analysis \citep{simonyan_deep_2014} and layer-wise relevance propagation \citep{bach_pixel-wise_2015}. These methods allow a visualization in terms of a heatmap the size of the input image. These algorithms have in common that the image only needs to be processed twice: the image data is propagated forward through the network and gradients of the image-level predictions, or similar quantities, are propagated backwards. The simplest solution \citep{simonyan_deep_2014}, for instance, computes the partial derivative of the image-level predictions with respect to the value of each pixel: the backpropagated quantities are partial derivatives of the image-level predictions. The most advanced solution \citep{bach_pixel-wise_2015} forces the backpropagated quantities to be preserved between neurons of two adjacent layers. A detailed comparison can be found in \citet{samek_evaluating_2016}.

For the joint detection of referable DR and DR lesions, we need a solution which can deal with multiple occurrences of the same lesion, unlike \citet{zeiler_visualizing_2014}'s solution, and which can deal with small lesions like microaneurysms, unlike \citet{worrall_automated_2016}'s solution. The above pixel-level visualization techniques are more relevant to our task. However, we will show that the heatmaps they produce contain artifacts caused by the architecture of ConvNets. We propose to reduce those artifacts through a joint optimization of the ConvNet predictions and of the produced heatmaps. Among those three algorithms, sensitivity analysis \citep{simonyan_deep_2014} is the only criterion that can be easily differentiated, and which is therefore compatible with the proposed optimization: we decided to base our solution on this criterion. To go from visualizations designed to help understand what ConvNets are learning to visualizations useful for computer-aided diagnosis, the quality of the produced heatmaps needs to be improved, as presented hereafter.

\section{Heatmap Generation}
\label{sec:HeatmapGeneration}

\subsection{Notations}

Let $L$ denote the number of layers in a ConvNet. Let $D^{(l)}$, $l=0, ..., L$, denote the data flowing from layer $l$ to layer $l+1$: $D^{(0)}$ denotes the input data, $D^{(l)}$, $1 \leq l \leq L-1$, is composed of activation maps and $D^{(L)}$ contains the image-level predictions. For faster computations, ConvNets usually process multiple images simultaneously, so $D^{(0)}$ is generally a mini-batch of $N$ images. $D^{(l)}$ is organized as a fourth-order tensor with dimensions $N\times W_{l}\times H_{l}\times C_{l}$, where $W_{l}\times H_{l}$ is the size of the activation maps produced by layer $l$ (or the size of the input images if $l=0$) and $C_{l}$ is the number of activation maps per image (or the number of color channels if $l = 0$). In dense layers, such as the prediction layer, each map contains a single value ($W_{l} = H_{l} = 1$): in that case, $D^{(l)}_{n,1,1,c}$ is written $D^{(l)}_{n,c}$ for short.

\subsection{Sensitivity Criterion}

The sensitivity criterion assesses the contribution of one color channel of one pixel, namely $D^{(0)}_{n,x,y,c}$, to the ConvNet's prediction that the image belongs to some class $d$, namely $D^{(L)}_{n,d}$. This criterion is defined as the absolute value of the partial derivative of $D^{(L)}_{n,d}$ with respect to $D^{(0)}_{n,x,y,c}$, which can be computed according to the chain rule of derivation:
\begin{equation}
  \label{eq:chainRule1}
  \frac{\partial D^{(L)}_{n,d}}{\partial D^{(l-1)}} = \frac{\partial D^{(L)}_{n,d}}{\partial D^{(l)}}\frac{\partial D^{(l)}}{\partial D^{(l-1)}}, \;\;l = L-1, ..., 1.
\end{equation}
Here, we focus on the `referable DR' class ($d = $`+'). If we denote by $f_n: \mathbb{R}^{N\times W_0\times H_0\times C_0} \rightarrow \mathbb{R}$ the ConvNet's prediction that image $n$ belongs to the `referable DR' class, the overall contribution $\omega_{n,x,y}$ of pixel $D^{(0)}_{n,x,y}$ can be summarized as follows \citep{simonyan_deep_2014}:
\begin{equation}
  \omega_{n,x,y} = \left\| \left( \frac{\partial D^{(L)}_{n,+}}{\partial D^{(0)}_{n,x,y,c}} \right)_{c \in \{r, g, b\} } \right\|_q = \left\| \left( \frac{\partial f_n(D^{(0)})}{\partial D^{(0)}_{n,x,y,c}} \right)_{c \in \{r, g, b\} } \right\|_q  \;\;,
  \label{eq:sensitivityCriterion}
\end{equation}
where $\|.\|_q, q\in \mathbb{N}$, denotes the $q$-norm; \citet{simonyan_deep_2014} used $q = \infty$.

\subsection{Interpretation}

The sensitivity criterion indicates which local changes would modify the network predictions. In the context of DR screening, this can be interpreted as follows. Let us assume that a fundus image is from a referable DR patient and classified as such by the ConvNet. In that case, any lesion should be associated with nonzero $\omega$ values, as removing the lesion might downgrade the diagnosis and enhancing the lesion would consolidate it. Now, let us assume that the image is not from a referable DR patient and classified as such by the ConvNet. In that case, subtle microaneurysms in mild nonproliferative DR patients, which are not referable yet, should be associated with nonzero $\omega$ values. Indeed, more pronounced microaneurysms would possibly upgrade the diagnosis to referable DR. So, in both cases, lesions should be detected by the $\omega$ criterion.

Although this criterion has interesting features for lesion detection, it also has a few drawbacks: two types of artifacts may appear, so we had to modify this criterion accordingly.

\subsection{Disguise Artifacts}

The first limitation of the sensitivity criterion is that it does not reveal directly whether a pixel contains evidence for or against the prediction made by a ConvNet: it simply gives, for every pixel, a direction in RGB-space in which the prediction increases or decreases \citep{samek_evaluating_2016}. In particular, nonzero $\omega$ values may also be associated with lesion confounders: dust on the camera's lens resembling microaneurysms, specular reflections resembling exudates or cotton-wool spots, etc. Indeed, modifying a confounder could make it resemble a true lesion even more. Typically, changing one or two color channels only would give it a more compatible color. Nonzero $\omega$ values may also be associated with healthy tissue surrounding a lesion: changing its color could make the lesion look bigger. We will refer to these phenomena as `disguise artifacts'.

\subsection{Hue-Constrained Sensitivity Criterion}

The solution we propose to reduce those artifacts is to constrain the allowed local changes. Specifically, we force these changes to preserve the hue, so that sensitivity analysis essentially focuses on pattern enhancements or attenuations. Hue preservation is ensured by forcing all three color components of a pixel to be multiplied by the same factor.  So, instead of computing the sensitivity of each color component independently and combining them afterwards (following Equation (\ref{eq:sensitivityCriterion})), a single sensitivity value $\pi_{n,x,y}$ is computed per pixel in a single operation, as described hereafter. Given the input tensor $D^{(0)}$ with dimensions $N\times W_0\times H_0\times C_0$, a binary tensor $m$ with dimensions $N\times W_0\times H_0\times 1$ is defined. The proposed hue-constrained sensitivity criterion is given by:
\begin{equation}
  \pi_{n,x,y} = \left| \frac{\partial f_n\left( m \circ D^{(0)} \right)}{\partial m_{n,x,y}} \right|  \;\;,
  \label{eq:hueConstrainedSensitivityCriterion}
\end{equation}
where tensor $m$ is filled with ones and where `$\circ$' denotes the entrywise tensor multiplication, which implies that $m \circ D^{(0)} = D^{(0)}$. Following the usual convention, the fact that the fourth dimension of $m$ is 1 implies that all color components of a pixel in $D^{(0)}$ are multiplied by the same tensor element in $m$, which ensures the desired hue preservation property.

\subsection{Drafting Artifacts}

\begin{figure*}[t]
  \begin{center}
      \includegraphics[width=0.73\textwidth]{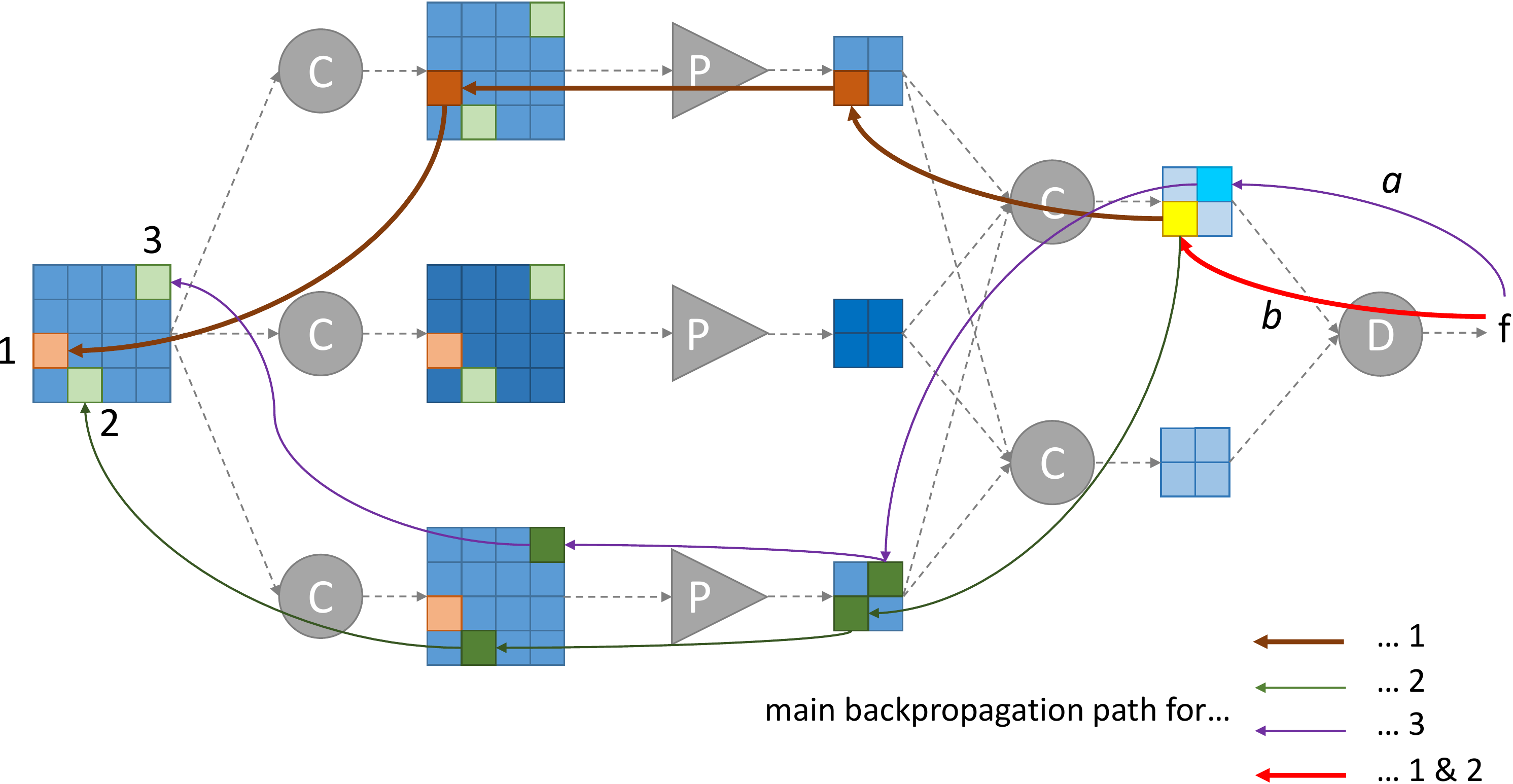}
  \end{center}
  \caption{Illustrating drafting artifacts: one lesion (`1', in red) and two confounders (`2' and `3', in green) are processed by a simplistic ConvNet. Convolution operations (stride = 1) and max-pooling operations (size = 2 $\times$ 2, stride = 2) are indicated by letters `C' and `P', respectively; dense layers are indicated by letter `D'. Colors in intermediate feature maps represent the contribution of each input pattern (red + green $\rightarrow$ yellow, green + blue $\rightarrow$ cyan). Because pattern `1' is a lesion, partial derivatives associated with edges along its backpropagation paths (such as edge `b') tend to be larger (compared to edge `a' in particular). Because they are neighbors, patterns `1' and `2' share the first edge (`b') along their main backpropagation path. So, even though patterns `2' and `3' are identical, the sensitivity of output `f' with respect to `3', obtained by multiplying partial derivatives along its backpropagation paths, is smaller than the sensitivity with respect to `2', which benefits from a drafting effect from `1'.}
  \label{fig:draftingArtifacts}
\end{figure*}

The second and most important limitation of the sensitivity criterion is that confounders in the vicinity of true lesions tend to be amplified. This effect, illustrated in Fig. \ref{fig:draftingArtifacts}, is due to down-sampling, which occurs in pooling or convolutional layers with a stride greater than one (see \ref{app:PopularConvNetOperators}). Indeed, according to the chain rule of derivation [see Equation (\ref{eq:chainRule1})], the gradient of $f_n(D^{(0)})$ with respect to $D^{(0)}$ is obtained by multiplying the following partial derivative tensors: $\frac{\partial D^{(1)}}{\partial D^{(0)}}$, ..., $\frac{\partial D^{(L)}}{\partial D^{(L-1)}}$. Because of down-sampling, these partial derivative tensors are of decreasing sizes. As a result, a true lesion and confounders in its vicinity share common terms in the expression of their influence on $f_n(D^{(0)})$. These terms tend to be large because of the true lesion, so the perceived influence of the confounders is artificially boosted. We will refer to those amplified false alarms as `drafting artifacts'. It should be noted that those drafting artifacts occur in all algorithms based on backpropagation, including the deconvolution method \citep{zeiler_visualizing_2014} and layer-wise relevance propagation \citep{bach_pixel-wise_2015}.

\subsection{Reducing Drafting Artifacts}

A brute-force solution for reducing those artifacts would be to 1) compute the $\pi$ tensor according to Equation (\ref{eq:hueConstrainedSensitivityCriterion}), 2) record the maximal $\pi_{n,x^*,y^*}$ values and 3) set the corresponding $m_{n,x^*,y^*}$ values to zero. Then, the $\pi$ tensor should be computed again using the modified $m$ tensor, in order to record the next largest $\pi_{n,x^{**},y^{**}}$ values, without the drafting influence of the $(n,x^*,y^*)$ pixels. And so on until the influence of each pixel has been recorded independently from its more influential neighbors. However, the complexity of this solution clearly is prohibitive. Instead, we propose an indirect solution which reduces drafting artifacts while training the ConvNet, so that we do not have to deal with them explicitly afterwards.

\section{Heatmap Optimization}
\label{sec:HeatmapOptimization}

\subsection{Training a ConvNet with the Backpropagation Method}

The parameters of a ConvNet (namely weights and biases --- see \ref{app:PopularConvNetOperators}) are generally optimized with the backpropagation method. This method progressively calculates the gradient of a loss function $\mathcal L_L$ with respect to each parameter tensor $\rho$, using the chain rule of derivation:
\begin{eqnarray}
  \label{eq:chainRule1Bis}
  \frac{\partial \mathcal L_L}{\partial D^{(l-1)}} &=& \frac{\partial \mathcal L_L}{\partial D^{(l)}}\frac{\partial D^{(l)}}{\partial D^{(l-1)}}  \;\;, \\
  \frac{\partial \mathcal L_L}{\partial \rho} &=& \frac{\partial \mathcal L_L}{\partial D^{(l)}}\frac{\partial D^{(l)}}{\partial \rho}  \;\;.
  \label{eq:chainRule2}
\end{eqnarray}
Those gradients are then used by an optimizer to update the parameters, in an attempt to minimize the loss function. Loss function $\mathcal L_L$ typically quantifies the classification or regression error, based on image-level predictions computed in $D^{(L)}$. To improve generalization, regularization terms are usually added to the loss function: they typically aim at minimizing the L1- or L2-norm of the filter weights.

\subsection{Sparsity-Enhanced Sensitivity Criterion}

In order to reduce drafting artifacts, we propose to include an additional regularization term $\mathcal L_0$ to the total loss function $\mathcal L$. The aim of $\mathcal L_0$ is to maximize the sparsity of $\omega$ or $\pi$. By forcing the ConvNet to reduce the number of nonzero pixels in $\omega$ or $\pi$, while maintaining its classification or regression accuracy, the ConvNet has to modify its parameters in such a way that true lesions and confounders in their vicinity share as little large terms as possible in the expression of their influence on $f_n(D^{(0)})$. In other words, the ConvNet is forced to build more discriminative filters: filters that better separate true lesions from confounders. Following \citet{tibshirani_regression_1996}, the sparsity of $\omega$ or $\pi$ is maximized through L1-norm minimization, rather than L0-norm minimization, which is NP-hard.

Because those heatmaps depend on backpropagated quantities, the network parameters cannot be optimized using the usual backpropagation method, so a different ConvNet training method had to be proposed: while standard training algorithms propagate image intensities through the ConvNet and backpropagate the gradients of the optimization criterion, the proposed training algorithm involves a third pass on the ConvNet to propagate second-order derivatives forward. This new training procedure can be obtained through simple adaptations of deep learning libraries.

\subsection{Backward-Forward Propagation Method}
\label{sec:BackwardForwardPropagationMethod}

\begin{figure}
  \begin{center}
      \includegraphics[width=0.8\textwidth]{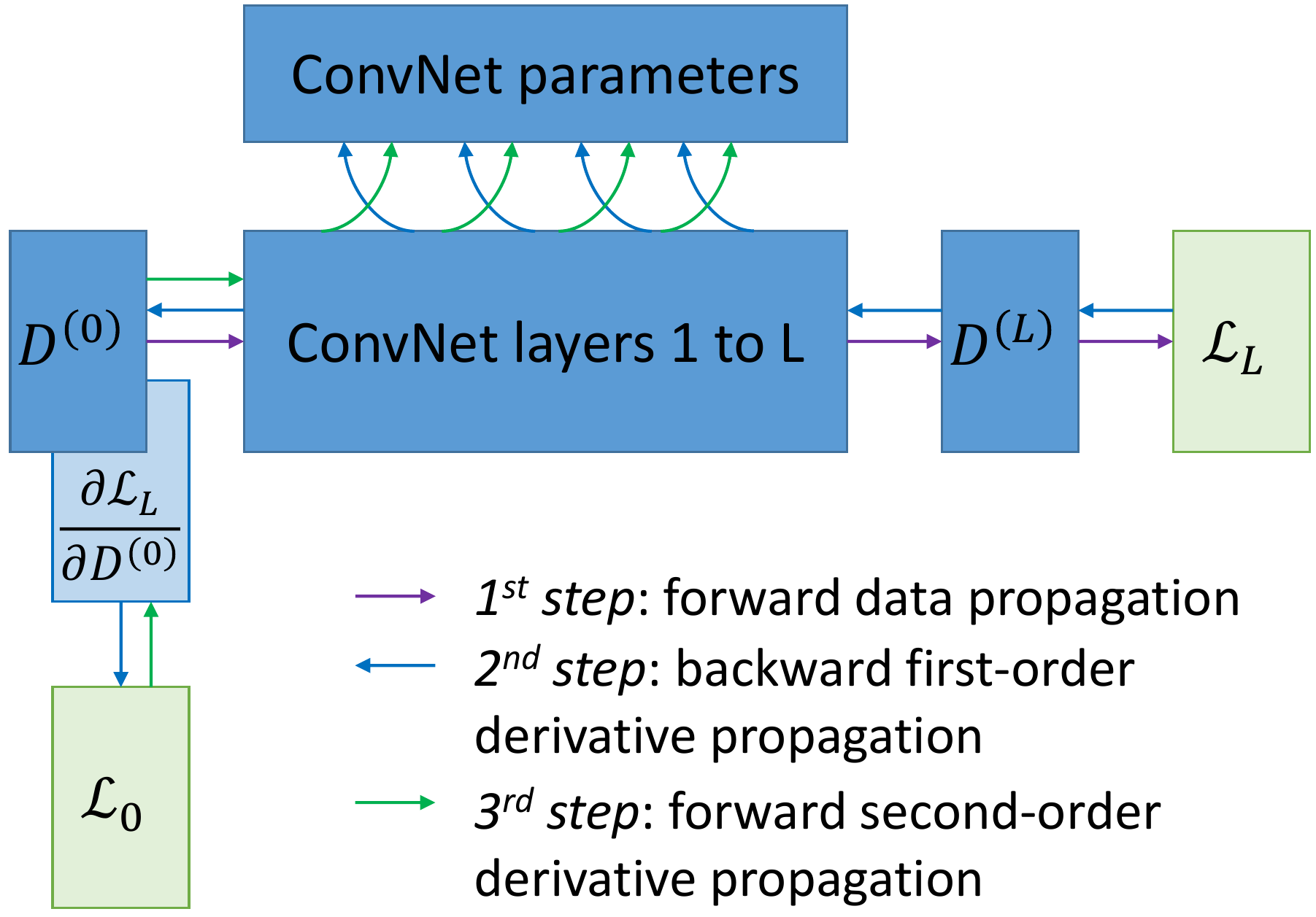}
  \end{center}
  \caption{Backward-forward propagation method.}
  \label{fig:backwardForward}
\end{figure}

We introduce a novel method for optimizing the parameters of a ConvNet when the loss function to minimize is of the form:
\begin{equation}
  \label{eq:loss}
  \left\lbrace
  \begin{array}{rcl}
    \mathcal L &=& \mathcal L_L + \mathcal L_0 \\
    \mathcal L_0 &=& g \left( \displaystyle\frac{\partial \mathcal L_L}{\partial D^{(0)}} \right)
  \end{array}
  \right. .
\end{equation}
In order to enhance the sparsity of $\omega$ maps, function $g$ is set to the L1-norm, multiplied by some factor $\nu$. The sparsity of $\pi$ maps is enhanced similarly: in this case, the role of the input data tensor is played by tensor $m$ [see Equation (\ref{eq:hueConstrainedSensitivityCriterion})].

The proposed algorithm, illustrated in Fig. \ref{fig:backwardForward}, updates each parameter tensor $\rho$ during one training iteration, as follows:
\begin{enumerate}
  \item The data is propagated forward through the network, from $D^{(0)}$ to $D^{(L)}$, in order to compute loss function $\mathcal L_L$.
  \item The gradient of $\mathcal L_L$ is propagated backward through the network, from $D^{(L)}$ to $D^{(0)}$, using Equations (\ref{eq:chainRule1Bis}) and (\ref{eq:chainRule2}). The goal is to compute $\tfrac{\partial \mathcal L_L}{\partial \rho}$, for each parameter tensor $\rho$, and also to compute $\mathcal L_0$.
  \item The gradient of $\mathcal L_0$ is propagated forward through the first-order derivative of the network, from $D^{(0)}$ to $D^{(L)}$, in order to compute $\tfrac{\partial \mathcal L_0}{\partial \rho}$, for each parameter tensor $\rho$.
  \item Each parameter tensor $\rho$ is updated proportionally to $\tfrac{\partial \mathcal L_L}{\partial \rho} + \tfrac{\partial \mathcal L_0}{\partial \rho}$.
\end{enumerate}
The proposed backward-forward propagation method can be implemented simply using deep learning libraries with built-in automatic differentiation, such as TensorFlow\footnote{\url{https://www.tensorflow.org} --- one line implementation of the $\mathcal L_0$ cost function, if $g$ is the L1-norm: \url{tf.add_to_collection('losses', tf.reduce_sum(tf.abs(tf.gradients(L_L, [D_l])[0])))}}. The main difficulty is to provide a forward second-order derivative function for each operator in the network, in order to perform step 3, while deep learning libraries only provide backward first-order derivatives. \ref{app:ForwardSecondOrderDerivativesPopularConvNetOperators} provides the forward second-order derivatives for operators used in the following experiments.

The proposed heatmap generation and optimization solution is now evaluated in the context of diabetic retinopathy screening, using ConvNets specifically designed for this task.

\section{Experiments}
\label{sec:Experiments}

\subsection{Baseline ConvNet}

This section introduces the ConvNets that we used in our experiments. These ConvNets produce predictions at the image level; based on modifications described in sections \ref{sec:HeatmapGeneration} and \ref{sec:HeatmapOptimization}, they also produce predictions at the pixel level. Successful solutions based on ConvNets were submitted to the 2015 Diabetic Retinopathy competition and the source code of the best solutions is publicly available. Rather than reinventing the wheel, we used the best of these solutions to set up our baseline ConvNets.

\subsubsection{Image Preprocessing and Data Augmentation}

Image preprocessing was adapted from the min-pooling solution,\footnote{\url{https://www.kaggle.com/c/diabetic-retinopathy-detection/forums/t/15801/competition-report-min-pooling-and-thank-you}} by B. Graham, which ranked first in the Kaggle Diabetic Retinopathy competition. Let $I$ denote the input image. The width of the camera's field of view in $I$ is estimated and $I$ is resized to normalize that width to 512 pixels. The background $I_b$ of the resulting $I_r$ image is then estimated by a large Gaussian filter in each color channel (standard deviation: 8.5 pixels). A normalized image is defined as $I_n=4(I_r-I_b)$. Finally, because the camera's field of view usually contains illumination artifacts around its edges, the field of view is eroded by 5\% in $I_n$. Following all the top ranking solutions in the competition, data augmentation is performed during training. Before feeding a preprocessed image to the network, the image is randomly rotated (range: $[0^{\circ}, 360^{\circ}]$), translated (range: [-10 px, 10 px]), scaled (range: $[85 \%, 115 \%]$), horizontally flipped and its contrast is modified (multiplicative factor range: $[60 \%, 167 \%]$); different transformation parameters are generated at each epoch. The resulting image is resized and cropped to 448 $\times$ 448 pixels.

\subsubsection{Network Structures}

The network structures used in this study were adapted from the o\_O solution,\footnote{\url{https://www.kaggle.com/c/diabetic-retinopathy-detection/forums/t/15617/team-o-o-solution-summary}} by M. Antony and S. Br\"uggemann, which ranked second in the Kaggle Diabetic Retinopathy competition. This solution was selected since it relies on networks composed exclusively of basic processing units implemented in all deep learning libraries. This property does not apply to the min-pooling solution, in particular, which relies on specialized operations such as fractional max pooling \citep{graham_fractional_2014}.

The o\_O solution relies on two networks, namely `net A' and `net B', applied to images of size 448 $\times$ 448 pixels. Their structure is described in table \ref{tab:structure}. It also relies on two sub-networks of `net A' and `net B' applied to smaller images (224 $\times$ 224 pixels and 112 $\times$ 112 pixels). All convolutional and dense layers use untied biases and leaky rectifiers as activation functions (see \ref{app:PopularConvNetOperators}). The last dense layer with a single unit is used for regression, to predict the image label. The designers of o\_O noticed that `net B' alone works almost as well as the ensemble, so we studied `net B' in more detail.

\begin{table*}
  \caption{Network Structure (see \ref{app:PopularConvNetOperators}) --- ConvNets from the o\_O solution.}
  \begin{center}
  {
    \footnotesize
    \begin{tabular}{|r||c|c|c|c|c|c|c|c|}
      \cline{4-9}
      \multicolumn{3}{c|}{} & \multicolumn{3}{|c|}{net `A'}                 & \multicolumn{3}{|c|}{net `B'} \\ 
      \hline
      \multirow{2}{*}{id} & layer & activation & window & window & output       & window & window & output        \\
                                      & type  & maps        & size       & stride    & tensor size & size       & stride    & tensor size \\
      \hline
      \hline
      1   & Input      & \multicolumn{3}{|c|}{} & 448 x 448 & \multicolumn{2}{|c|}{} & 448 x 448 \\
      \hline
      2   & Conv      & 32                  & 5 x 5      & 2        & 224 x 224  & 4 x 4      & 2        & 224 x 224\\
      3   & Conv      & 32                  & 3 x 3      & 1        & 224 x 224  & 4 x 4      & 1        & 225 x 225\\
      4   & MaxPool & $\emptyset$ & 3 x 3      & 2        & 111 x 111  & 3 x 3      & 2        & 112 x 112\\
      \hline
      5   & Conv      & 64                  & 3 x 3      & 2        & 56 x 56 & 4 x 4      & 2        & 56 x 56 \\
      6   & Conv      & 64                  & 3 x 3      & 1        & 56 x 56 & 4 x 4      & 1        & 57 x 57 \\
      7   & Conv      & 64                  & 3 x 3      & 1        & 56 x 56 & 4 x 4      & 1        & 56 x 56 \\
      8   & MaxPool & $\emptyset$ & 3 x 3      & 2        & 27 x 27  & 3 x 3      & 2        & 27 x 27 \\
      \hline
      9   & Conv      & 128                & 3 x 3      & 1        & 27 x 27 & 4 x 4      & 1        & 28 x 28 \\
      10 & Conv      & 128                & 3 x 3      & 1        & 27 x 27 & 4 x 4      & 1        & 27 x 27 \\
      11 & Conv      & 128                & 3 x 3      & 1        & 27 x 27 & 4 x 4      & 1        & 28 x 28 \\
      12 & MaxPool & $\emptyset$ & 3 x 3      & 2        & 13 x 13 & 3 x 3      & 2        & 13 x 13 \\
      \hline
      13 & Conv      & 256               & 3 x 3       & 1        & 13 x 13 & 4 x 4       & 1        & 14 x 14 \\
      14 & Conv      & 256               & 3 x 3       & 1        & 13 x 13 & 4 x 4       & 1        & 13 x 13 \\
      15 & Conv      & 256               & 3 x 3       & 1        & 13 x 13 & 4 x 4       & 1        & 14 x 14 \\
      16 & MaxPool & $\emptyset$ & 3 x 3      & 2        & 6 x 6     & 3 x 3      & 2        & 6 x 6 \\
      \hline
      17 & Conv      & 512               & 3 x 3      & 1        & 6 x 6 & 4 x 4      & 1        & 5 x 5 \\
      18 & Conv      & 512               & 3 x 3      & 1        & 6 x 6 & $\emptyset$ & $\emptyset$ & $\emptyset$ \\
      19 & RMSPool & $\emptyset$ & 3 x 3     & 3        & 2 x 2 & 3 x 3     & 2        & 2 x 2 \\
      \hline
      20 & Dropout & $\emptyset$ & \multicolumn{3}{|c|}{} & \multicolumn{3}{|c|}{} \\
      21 & Dense    & 1024 & \multicolumn{3}{|c|}{} & \multicolumn{3}{|c|}{} \\
      22 & Maxout  & 512   & \multicolumn{3}{|c|}{} & \multicolumn{3}{|c|}{} \\
      \cline{1-3}
      23 & Dropout & $\emptyset$ & \multicolumn{3}{|c|}{} & \multicolumn{3}{|c|}{} \\
      24 & Dense    & 1024 & \multicolumn{3}{|c|}{} & \multicolumn{3}{|c|}{} \\
      25 & Maxout  & 512   & \multicolumn{3}{|c|}{} & \multicolumn{3}{|c|}{} \\
      \cline{1-3}
      26 & Dense    & 1       & \multicolumn{3}{|c|}{} & \multicolumn{3}{|c|}{} \\
      \hline
    \end{tabular}
  }
  \end{center}
  \label{tab:structure}
\end{table*}

To show the generality of the approach, the popular AlexNet structure was also evaluated \citep{krizhevsky_imagenet_2012}. Unlike `net A' and `net B', AlexNet processes images of size 224 $\times$ 224 pixels, so images had to be downsampled by a factor of 2. Downsampling was done dynamically, using a mean pooling operator with a stride of 2 and a window size of 2 $\times$ 2 pixels (see \ref{app:PopularConvNetOperators}), in order to produce heatmaps with 448 $\times$ 448 pixels.

\subsubsection{Network Training}
\label{sec:BaselineNetworkTraining}

Following o\_O, networks are trained to minimize the mean squared error between image labels and predictions. Additionally, L2 regularization with factor 0.0005 is applied to filter weights in all convolutional and dense layers. We use very leaky rectifiers ($\alpha = 0.33$) instead of leaky rectifiers ($\alpha = 0.01$) in o\_O. This allows us to train all layers simultaneously, using the Adam optimizer \citep{kingma_adam:_2015}. Antony and Br\"uggemann tried a similar strategy, but with the optimizer they used, namely the Nesterov momentum optimizer \citep{nesterov_method_1983}, it did not work well. A learning rate of 0.0001 was used initially. Following common practice, we manually decreased the learning rate by a factor of 10 when performance in the validation set stopped increasing.

\subsection{Implementation Details}

The proposed algorithms were implemented in C++ and Python using OpenCV\footnote{\url{http://opencv.willowgarage.com/}} for image preprocessing and data augmentation, and TensorFlow for network training and inference. Forward second-order derivatives were implemented in Python when possible; that of MaxPool was implemented in C++. One GPU card was used: a GeForce GTX 1070 by Nvidia. Training and testing were performed using mini-batches of $N = 36$ images, in accordance with the memory capacity of the GPU card (7.92 GB). Each ConvNet was trained with 350,000 mini-batches, i.e. with 350,000 $N$ = 12.6 million images generated through data augmentation.

\subsection{Datasets}

Three datasets were used in this study: the `Kaggle Diabetic Retinopathy' dataset, used for training and testing at the image level, `DiaretDB1', for testing at the lesion level and at the image level and also for improving performance at the image level, and finally `e-ophtha', for testing at the image level.

\subsubsection{Kaggle Diabetic Retinopathy Dataset}

The first dataset consists of 88,702 color fundus photographs from 44,351 patients: one photograph per eye.\footnote{\url{https://www.kaggle.com/c/diabetic-retinopathy-detection/data}} Images were captured with various digital fundus cameras, in multiple primary care sites throughout California and elsewhere. Their definitions range from 433 x 289 pixels to 5184 x 3456 pixels (median definition: 3888 x 2592 pixels). Those images were then uploaded to EyePACS, a free platform for DR screening \citep{cuadros_eyepacs:_2009}. For each eye, DR severity was graded by a human reader according to the ETDRS scale \citep{wilkinson_proposed_2003}: `absence of DR', `mild non-proliferative DR (NPDR)', `moderate NPDR', `severe NPDR' and `proliferative DR (PDR)'. The dataset was split into a training set (35,126 images from 17,563 patients) and a test set (53,576 images from 26,788 patients): those two sets are referred to as `Kaggle-train' and `Kaggle-test', respectively. Networks were trained on 80 $\%$ of the Kaggle-train dataset (the first 28,100 images) and validated on the remaining 20 $\%$ (the last 7,024 images).

For the purpose of this study, about DR screening, severity grades were grouped into two categories: nonreferable DR (absence of DR or mild NPDR) versus referable DR (moderate NPDR or more). The prevalence of referable DR was 19.6 $\%$ in Kaggle-train and 19.2 $\%$ in Kaggle-test.

\subsubsection{DiaretDB1 Dataset}
\label{sec:DiaretDB1Dataset}

The second dataset consists of 89 color fundus photographs collected at the Kuopio University Hospital, in Finland \citep{kauppi_diaretdb1_2007}. Images were captured with the same fundus camera, a ZEISS FF450\textit{plus} digital camera with a 50 degree field-of-view. Images all have a definition of 1500 x 1152 pixels. Independent markings were obtained for each image from four medical experts. The experts were asked to manually delineate the areas containing microaneurysms (or `small red dots'), hemorrhages, hard exudates and cotton wool spots (or `soft exudates') and to report their confidence ($<$ 50 $\%$, $\geq$ 50 $\%$, 100 $\%$) for each segmented lesion. Based on these annotations, only five images in the dataset are considered normal: none of the experts suspect these images to contain any lesions.

Given a target lesion type, \citet{kauppi_diaretdb1_2007} proposed a standardized procedure to evaluate the performance of a lesion detector, at the image level, in the DiaretDB1 dataset. In this purpose, one probability map was constructed per image: this map was obtained by averaging, at the pixel level, confidences from all four experts for the target lesion type. If and only if this map contains at least one pixel with an average confidence level above 75 $\%$, then the image is considered to contain the target lesion. Based on this criterion, a receiver-operating characteristic (ROC) curve can be constructed for the lesion detector.

\subsubsection{e-ophtha Dataset}
\label{sec:EOphthaDataset}

The third dataset consists of 107,799 photographs from 25,702 examination records: generally two photographs per eye, i.e. four photographs per examination record. These photographs were collected in the OPHDIAT screening network in the Paris area \citep{Erginay2008}. Images were captured either with a CR-DGi retinograph (Canon, Tokyo) or with a TRC-NW6S (Topcon, Tokyo) retinograph. Their definitions range from 1440 $\times$ 960 to 2544 $\times$ 1696 pixels. Up to 27 contextual fields were included in each record. This includes 9 demographic information fields (age, gender, weight, etc.) and 18 information fields related to diabetes. Each examination record was analyzed later on by one ophthalmologist, out of 11 participating ophthalmologists, in Lariboisi\`ere Hospital (Paris). The ophthalmologist graded DR severity in both eyes. It should be noted that the association between photographs and eyes is unknown. Therefore, the task we address in this study is to detect whether or not the patient has referable DR in at least one of his or her eyes. The dataset was split into a training set of 12,849 examination records (`e-ophtha-train') and a test set of 12,853 records (`e-ophtha-test'), described in \citet{quellec_automatic_2016}.

\subsection{Visualization Artifacts}

\begin{figure*}
  \begin{center}
    \includegraphics[width=0.7\textwidth]{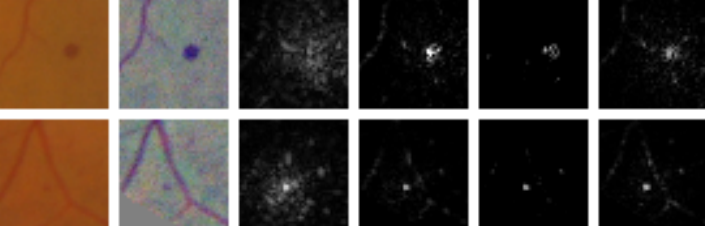}
  \end{center}
  \caption{Artifacts from various visualization algorithms using `net B'. From left to right: the original image, the preprocessed image, the original sensitivity map, the hue-constrained sensitivity map, the hue-constrained and sparsity-enhanced sensitivity map, and the layer-wise relevance propagation map.}
  \label{fig:visualization_artifacts}
\end{figure*}

Artifacts from various visualization algorithms are illustrated in Fig. \ref{fig:visualization_artifacts}. First, it can be seen that the original sensitivity criterion is inadequate to finely detect lesions. Sensitivity maps seem to indicate that if lesions grew in size, the diagnosis would be consolidated. The hue constraint prevents the lesion detections from growing. Second, it can be seen that, due to the drafting effect, blood vessels (lesion confounders) in the vicinity of lesions are detected, both in the hue-constrained sensitivity maps and in the layer-wise relevance propagation maps. The resulting false detections are not necessarily connected to the true detection, so they cannot be removed easily through image post-processing techniques: they have to be removed beforehand, hence the proposed approach.

\subsection{Image- and Pixel-Level Performance of ConvNets}

\begin{figure*}
  \begin{center}
    \begin{tabular}{c}
      \subfloat[Hue-constrained sensitivity criterion --- factor of the $\mathcal{L}_0$ cost function: $\nu = 0$]{
        \includegraphics[width=0.6\textwidth]{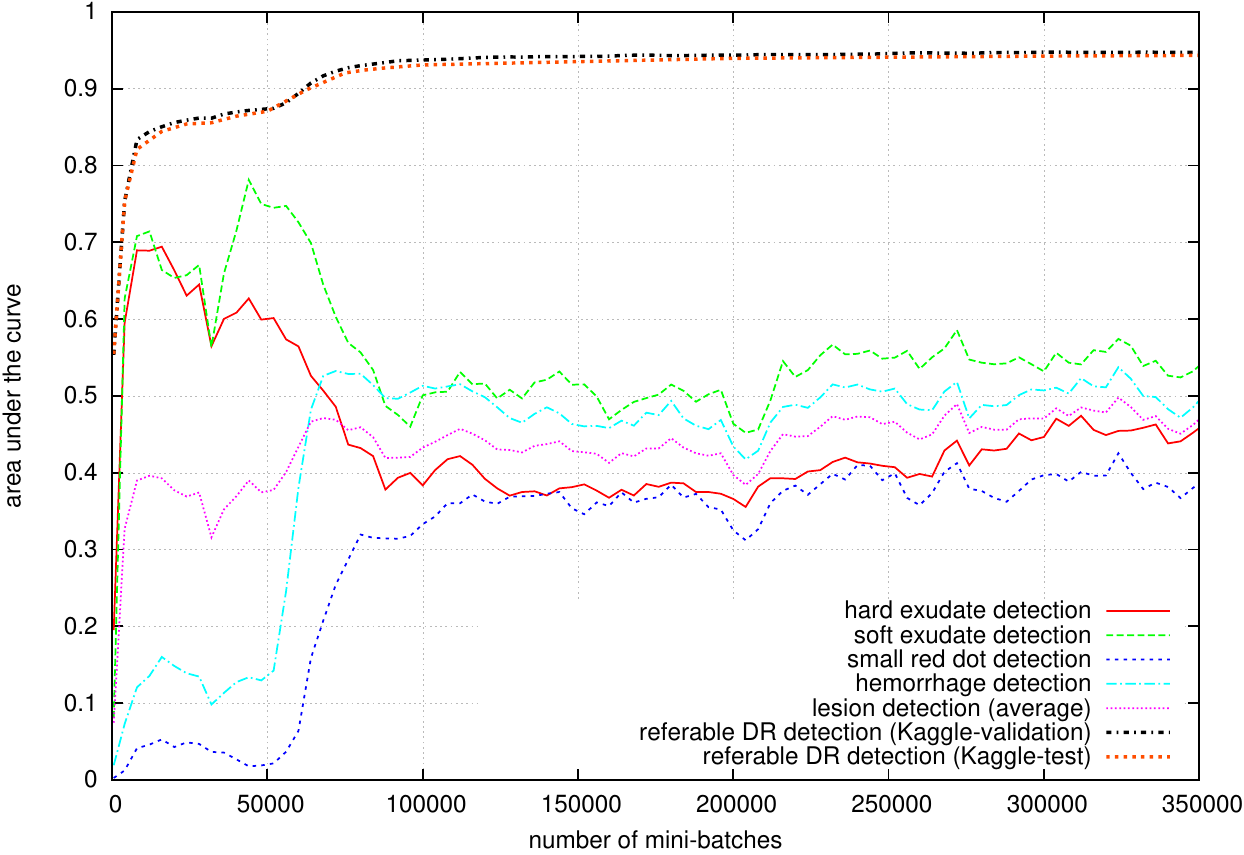}
      } \\
      \subfloat[Hue-constrained and sparsity-enhanced sensitivity criterion --- factor of the $\mathcal{L}_0$ cost function: $\nu = 10^{-3}$]{
        \includegraphics[width=0.6\textwidth]{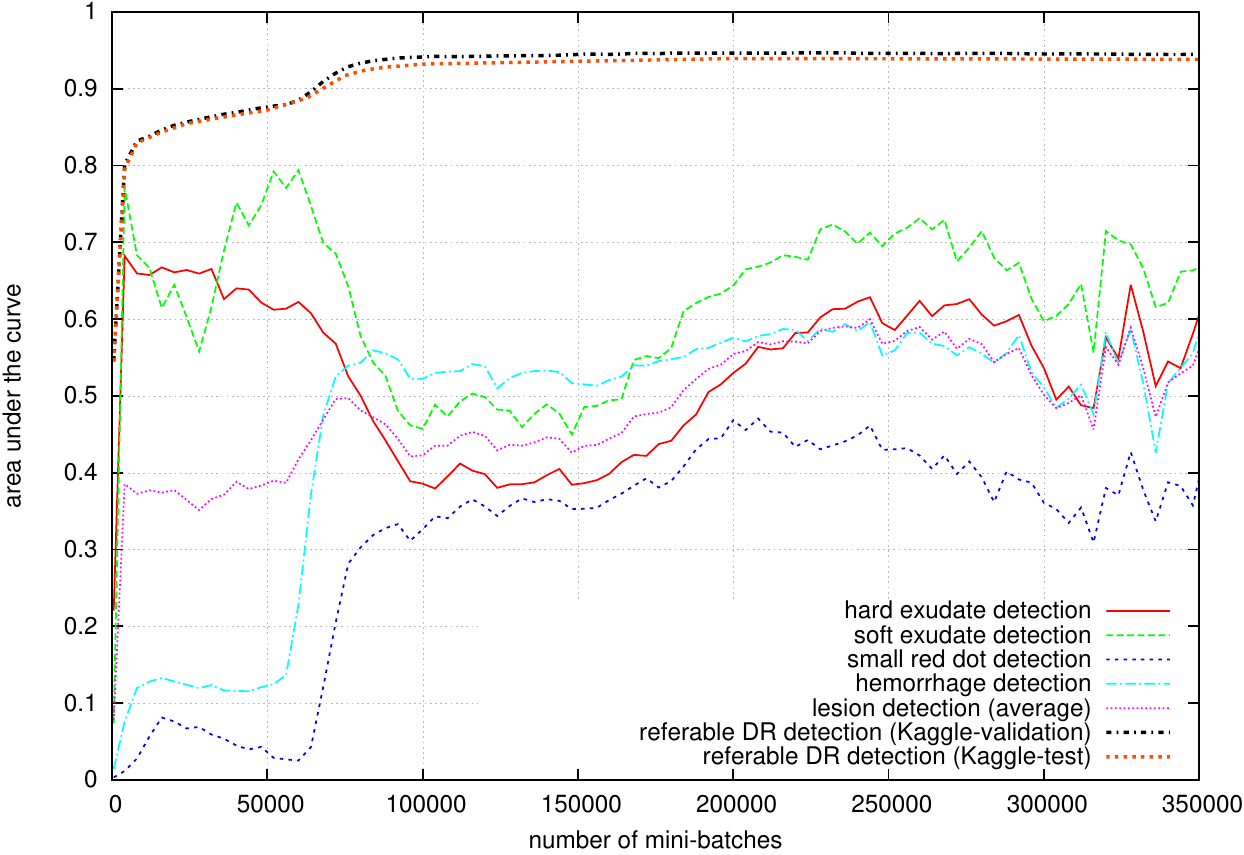}
      }
    \end{tabular}
  \end{center}
  \caption{ROC analysis for `net B' in Kaggle-validation and Kaggle-test --- FROC analysis in DiaretDB1.}
  \label{fig:validationPerformance}
\end{figure*}

Figure \ref{fig:validationPerformance} reports the performance of `net B' at different checkpoints stored during the training process. The hue-constrained sensitivity criterion is used, with or without sparsity maximization. Performance at the image level was assessed using a ROC analysis in the validation subset of Kaggle-train (`Kaggle-validation' for short), as well as in Kaggle-test: the area $A_z$ under the ROC curve is used as performance metric. Performance at the lesion level was assessed using a free-response ROC (FROC) analysis in the DiaretDB1 dataset. FROC curves are usually not bounded along the x-axis (the number false positives per image): we used as performance metric the area $A_z$ under the FROC curve for $0 \leq x \leq \mu$, divided by $\mu$ (with $\mu$ = 10). Performance is computed for each lesion type independently, and an average performance metric $\bar{A_z}$ is also computed. The influence of $\nu$, the factor of the $\mathcal{L}_0$ cost function, on the performance at the image level ($A_z$ in the Kaggle validation set) and at the lesion level ($\bar{A_z}$ in the DiaretDB1 dataset), is given in table \ref{tab:influenceNu}.
\begin{table}
  \caption{Influence of $\nu$, the factor of the $\mathcal{L}_0$ cost function,  on the performance of `net B' --- the retained solution is in bold.}
  \begin{center}
  {
    \footnotesize
    \begin{tabular}{c|cc}
    $\nu$ & image level $A_z$ & lesion level $\bar{A_z}$ \\
    \hline
    0                                   & 0.948 & 0.499 \\
    10\textsuperscript{-4} & 0.948 & 0.543 \\
    \textbf{10\textsuperscript{-3}} & \textbf{0.947} & \textbf{0.600} \\
    10\textsuperscript{-2} & 0.546 & 0.375 \\
    \end{tabular}
  }
  \end{center}
  \label{tab:influenceNu}
\end{table}

\begin{figure*}
  \begin{center}
    \begin{tabular}{cc}
      \subfloat[Hard exudates]{
        \includegraphics[width=0.35\textwidth]{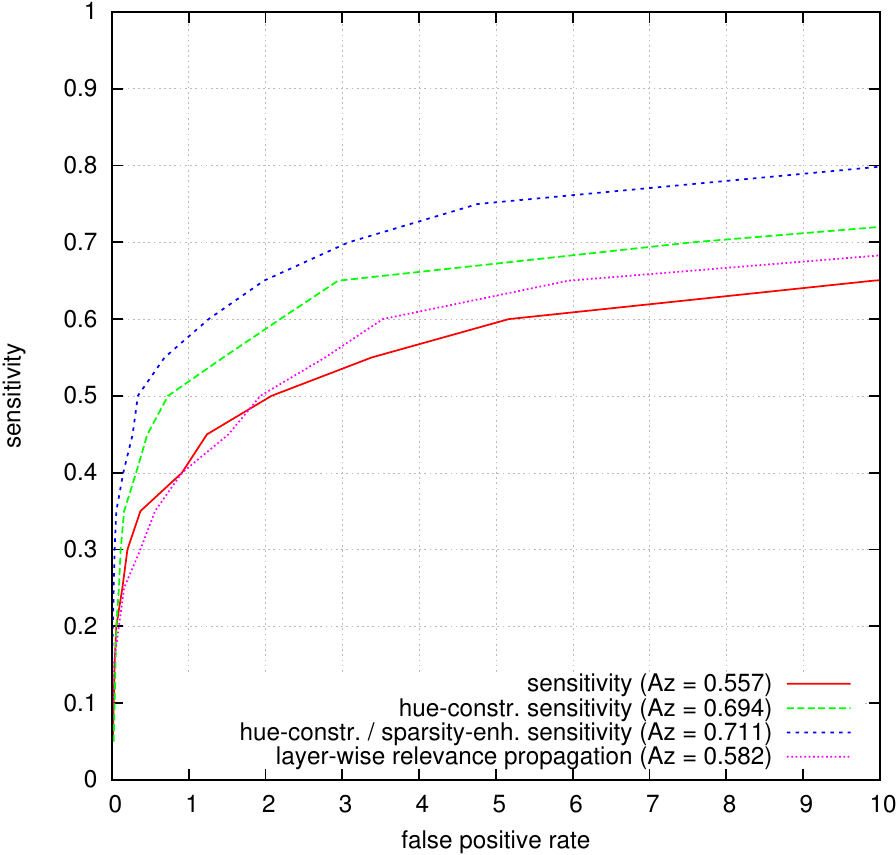}
      } &
      \subfloat[Soft exudates]{
        \includegraphics[width=0.35\textwidth]{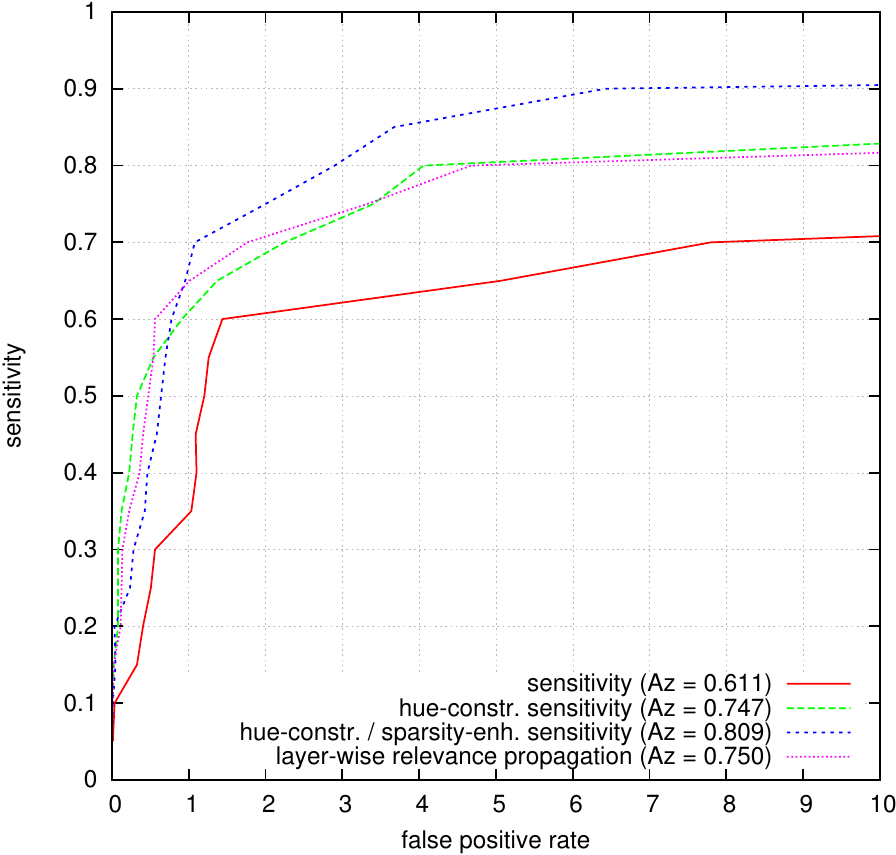}
      } \\
      \subfloat[Small red dots]{
        \includegraphics[width=0.35\textwidth]{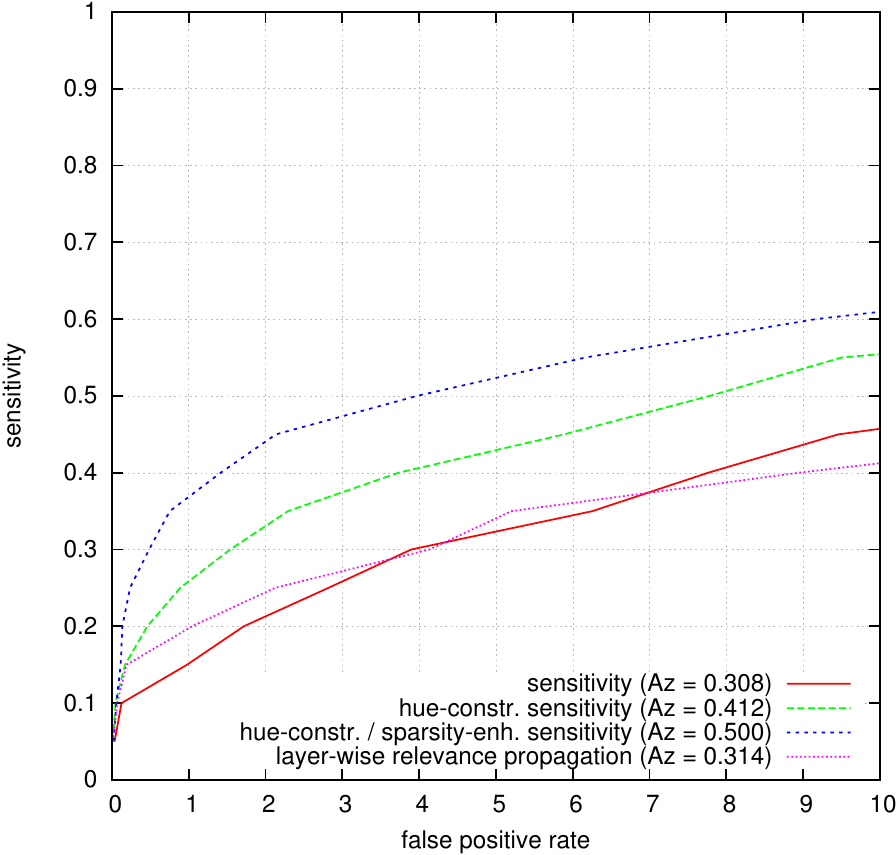}
      } &
      \subfloat[Hemorrhages]{
        \includegraphics[width=0.35\textwidth]{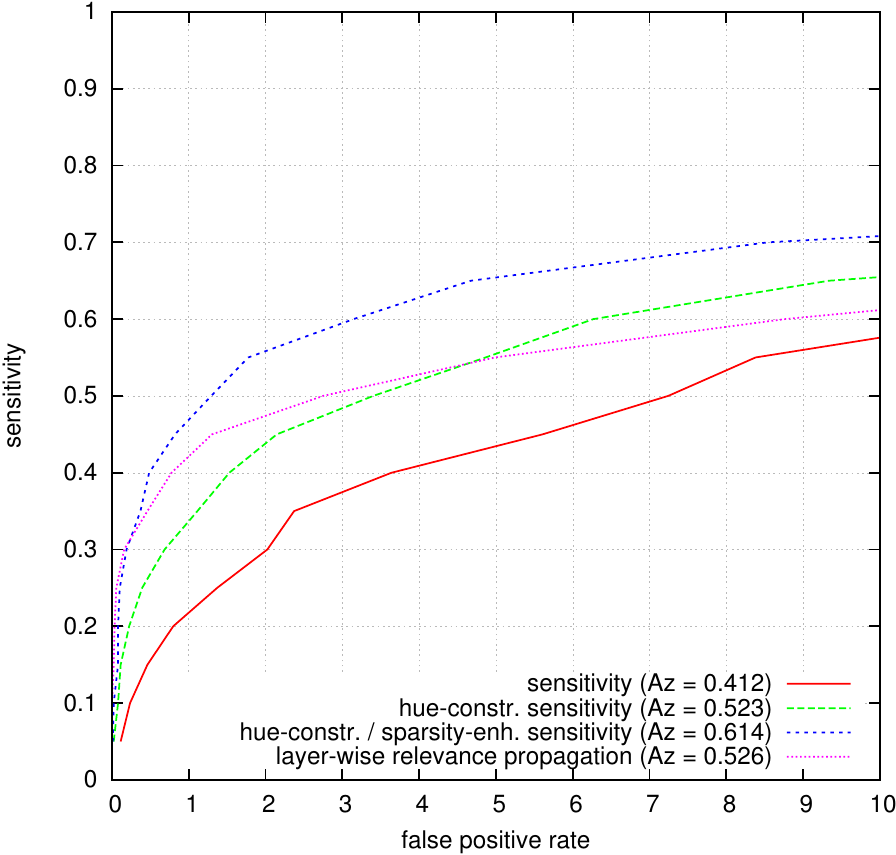}
      }
    \end{tabular}
  \end{center}
  \caption{Lesion detection performance in DiaretDB1 for `net B' at the lesion level.}
  \label{fig:diaretdb1_froc}
\end{figure*}

\begin{figure*}
  \begin{center}
    \begin{tabular}{cc}
      \subfloat[Hard exudates]{
        \includegraphics[width=0.35\textwidth]{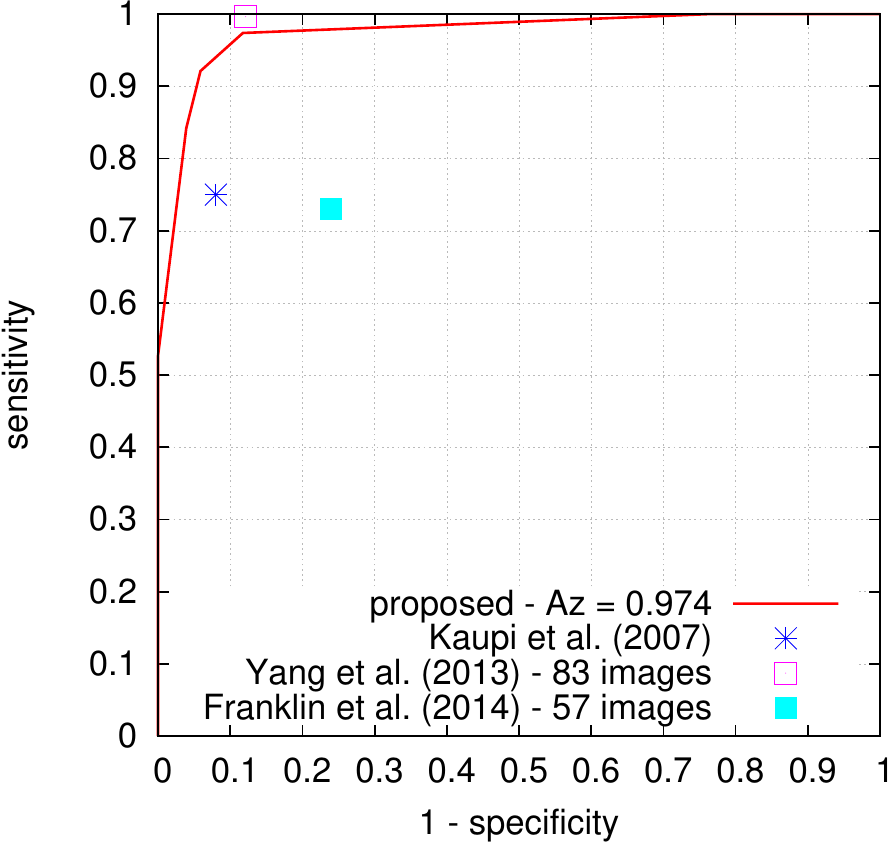}
      } &
      \subfloat[Soft exudates]{
        \includegraphics[width=0.35\textwidth]{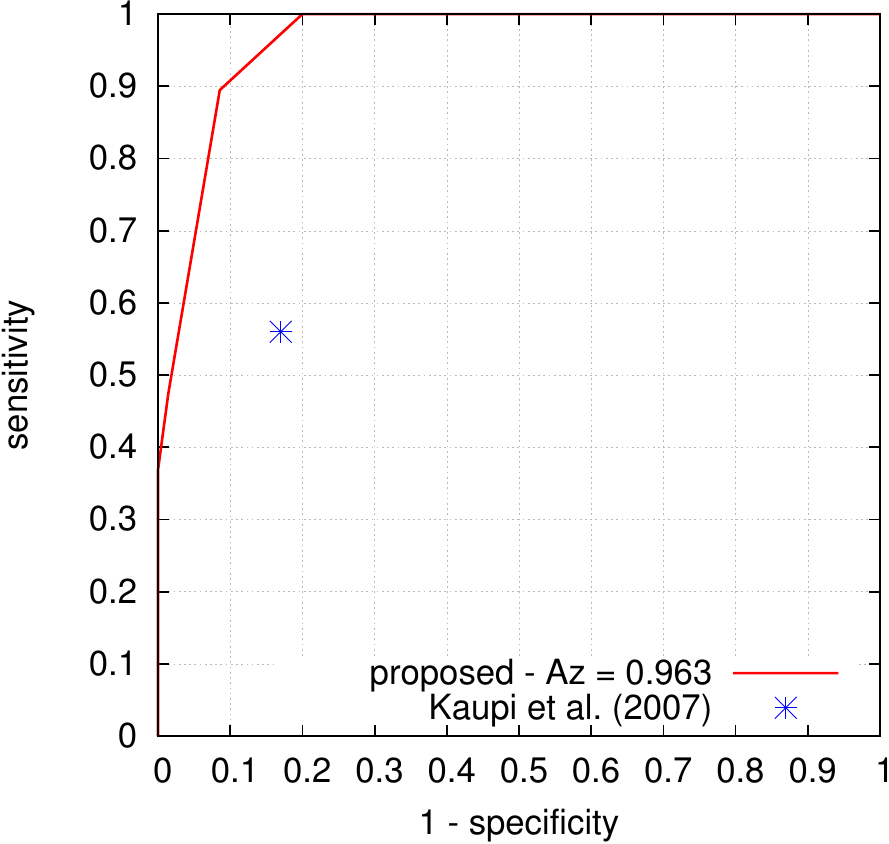}
      } \\
      \subfloat[Small red dots]{
        \includegraphics[width=0.35\textwidth]{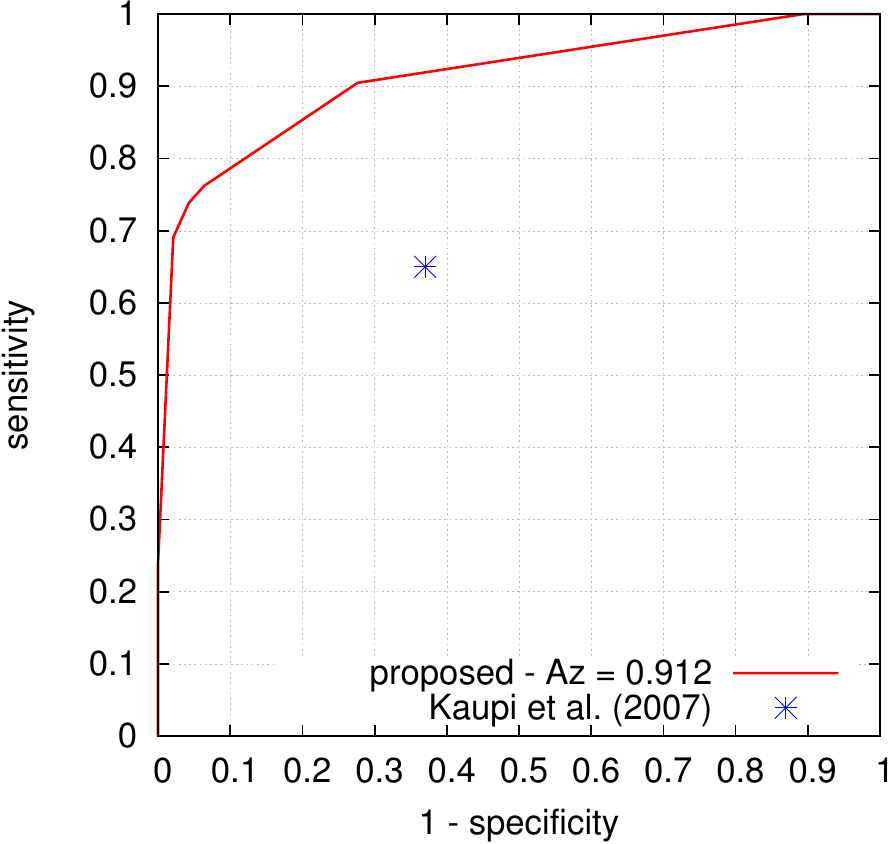}
      } &
      \subfloat[Hemorrhages]{
        \includegraphics[width=0.35\textwidth]{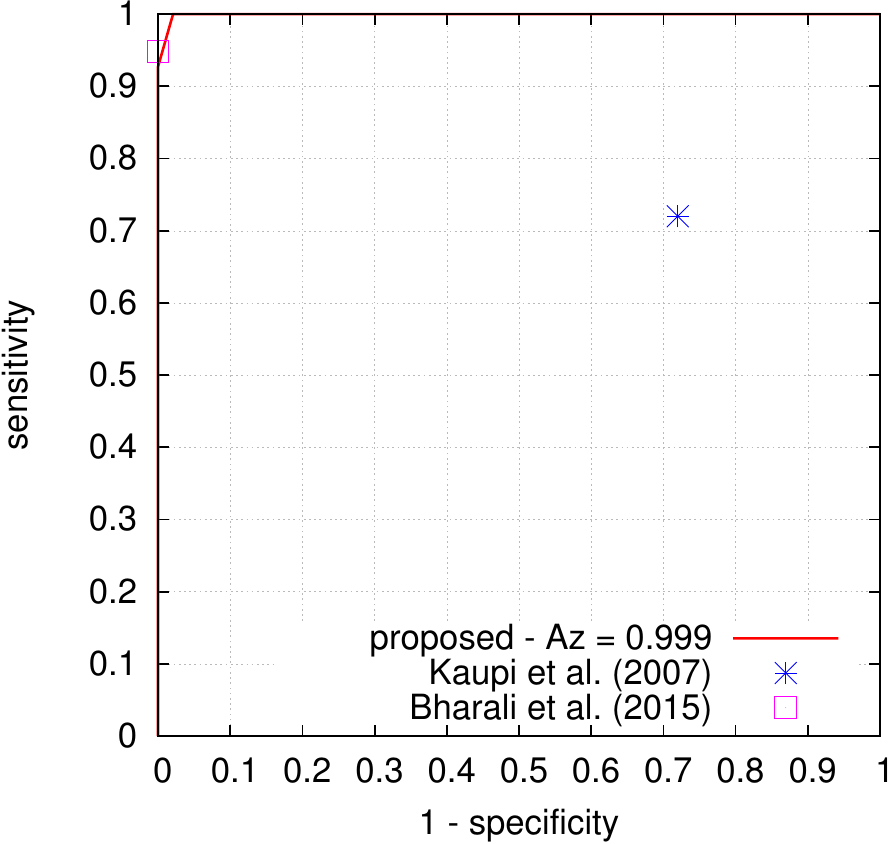}
      } \\
      \subfloat[Red lesions (hemorrhages and small red dots)]{
        \includegraphics[width=0.35\textwidth]{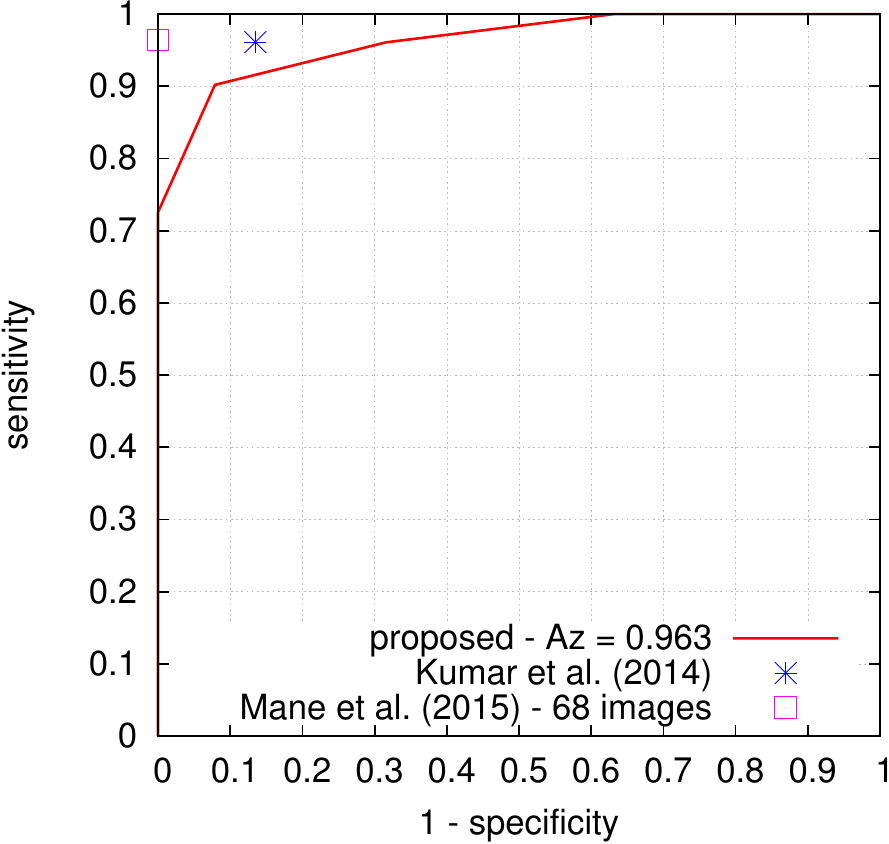}
      } &
      \subfloat[All lesions]{
        \includegraphics[width=0.35\textwidth]{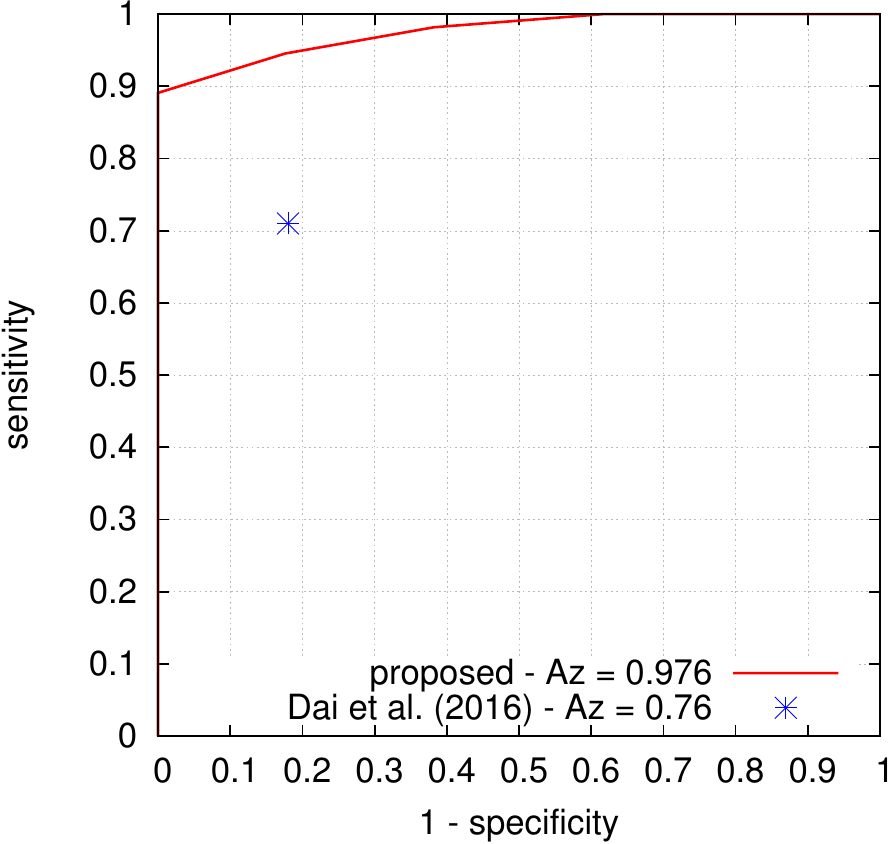}
      }
    \end{tabular}
  \end{center}
  \caption{Lesion detection performance for `net B' in DiaretDB1, at the image level, following the standardized procedure.}
  \label{fig:diaretdb1_roc}
\end{figure*}

To assess detection performance for each lesion type, a ten-fold cross-validation was performed. For each fold,
\begin{enumerate}
  \item the optimal checkpoint, as well as the optimal $\nu$ value when applicable ($\nu \in \{10^{-4}, 10^{-3}, 10^{-2}\}$), were found using 90\% of DiaretDB1 images,
  \item lesion probabilities were computed for the remaining 10\% using the optimal checkpoint and $\nu$ value.
\end{enumerate}
FROC curves are reported in Fig. \ref{fig:diaretdb1_froc} for `net B'; areas under the limited FROC curve for all three ConvNets are summarized in Table \ref{tab:diaretdb1_froc}. It appears that the number of false alarms is rather large, particularly in the case of microaneurysm detection. The reason is that human experts primarily segmented the most obvious lesions, while screening algorithms need to focus on the most subtle lesions as well. In other words, many true lesions are counted as false alarms. Of course, this comment also applies to competing automatic solutions. To show the value of our detections, the proposed solution was compared in Fig. \ref{fig:diaretdb1_roc} to results reported in the literature, following the DiaretDB1 standardized procedure (see section \ref{sec:DiaretDB1Dataset}). Most authors reported a single (sensitivity, specificity) pair: this is what we reported in Fig. \ref{fig:diaretdb1_roc}. Some authors reported ROC curves; in that case, we also reported a single (sensitivity, specificity) pair: the one closest to the (sensitivity = 1, specificity = 1) coordinate. Note that all competing solutions \citep{kauppi_diaretdb1_2007, yang_effective_2013, franklin_diagnosis_2014, kumar_automatic_2014, bharali_detection_2015, mane_detection_2015, dai_fundus_2016} are trained at the lesion level, while ours is trained (in Kaggle-train) at the image level.

\begin{table*}
  \caption{Lesion detection performance in DiaretDB1 at the lesion level --- area under the FROC curve for $0 \leq x \leq 10$, divided by 10. The best ConvNet/criterion couples are in bold.}
  \begin{center}
  {
    \footnotesize
    \begin{tabular}{c|c||c|c|c|c}
    \multirow{2}{*}{criterion} & \multirow{2}{*}{ConvNet} & hard         & soft          & small       & \multirow{2}{*}{hemorrhages} \\
                                               &                                            & exudates  & exudates & red dots &                                                    \\
    \hline
    \multirow{3}{*}{sensitivity} & net A     & 0.552 & 0.624 & 0.316 & 0.449 \\
                                                  & net B     & 0.557 & 0.611 & 0.308 & 0.412 \\
                                                  & AlexNet & 0.474 & 0.642 & 0.028 & 0.274 \\
    \hline
    hue-constrained                    & net A     & 0.677 & 0.754 & 0.393 & 0.533 \\
    sensitivity                              & net B     & 0.694 & 0.747 & 0.412 & 0.523 \\
                                                  & AlexNet & 0.516 & 0.742 & 0.032 & 0.300 \\
    \hline
    hue-constrained                    & net A     & \textbf{0.735} & 0.802 & 0.482 & 0.578 \\
    sparsity-enhanced                & net B     & 0.711 & \textbf{0.809} & \textbf{0.500} & \textbf{0.614} \\
    sensitivity                              & AlexNet & 0.618 & 0.783 & 0.032 & 0.320 \\
    \hline
    layer-wise                             & net A     & 0.600 & 0.743 & 0.315 & 0.519 \\
    relevance                              & net B     & 0.582 & 0.750 & 0.314 & 0.526 \\
    propagation                          & AlexNet & 0.499 & 0.747 & 0.027 & 0.274 \\
    \end{tabular}
  }
  \end{center}
  \label{tab:diaretdb1_froc}
\end{table*}

\subsection{Ensemble Learning}

As commonly done in machine learning competitions \citep{russakovsky_imagenet_2015}, an ensemble of ConvNets was used to boost performance. As less commonly done, all ConvNets in the ensemble originate from the same network, but with parameter values obtained at different checkpoints during the learning process. This way, training the ensemble is not significantly more computationally intensive than training a single ConvNet. As shown in Fig. \ref{fig:validationPerformance}, individual lesion types are not optimally detected after the same number of iterations. So, the idea was to export parameter values from the ConvNet when:
\begin{enumerate}
  \item hard exudates were optimally detected (network $\mathcal{N}_{HE}$ --- iteration 4,000 for `net B'),
  \item soft exudates were optimally detected (network $\mathcal{N}_{SE}$ --- iteration 60,000 for `net B'),
  \item small red dots were optimally detected (network $\mathcal{N}_{SRD}$ --- iteration 208,000 for `net B'),
  \item hemorrhages were optimally detected (network $\mathcal{N}_{H}$ --- iteration 244,000 for `net B'),
  \item lesions were optimally detected on average (network $\mathcal{N}_{AVG}$ --- iteration 328,000 for `net B'),
  \item referable DR was optimally detected (network $\mathcal{N}_{RDR}$ --- iteration 224,000 for `net B').
\end{enumerate}
In the Kaggle DR dataset, network predictions were computed for the current eye, but also for the contralateral eye, so 6 $\times$ 2 = 12 features were fed to the ensemble classifier per eye. In e-ophtha, network predictions were computed for all images in the current examination record. In regular examination records, consisting of four images, the four predictions computed at each checkpoint were stored in ascending order. In other records, the two lowest and the two largest predictions were stored. These 6 $\times$ 4 = 24 predictions were fed to the ensemble classifier, with up to 27 contextual features (see section \ref{sec:EOphthaDataset}). Following the min-pooling solution, a random forest was used to build the ensembles. These ensembles were trained in the full Kaggle-train and e-ophtha-train datasets, respectively.

\begin{figure}[t]
  \begin{center}
    \begin{tabular}{c}
        \includegraphics[width=0.9\textwidth]{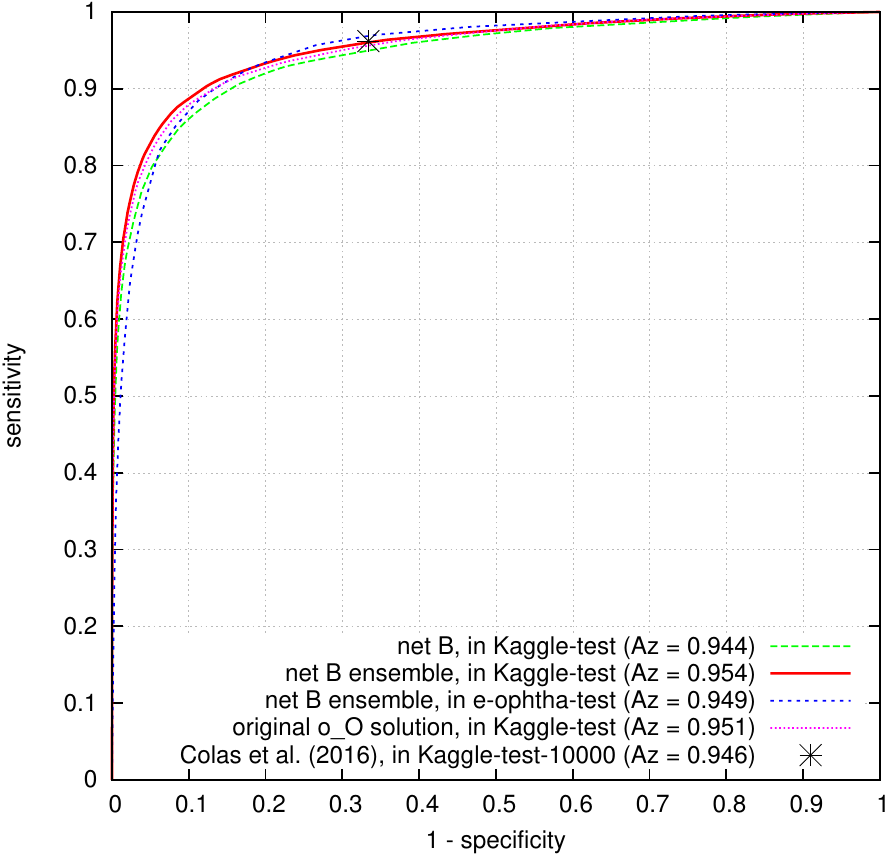}
    \end{tabular}
  \end{center}
  \caption{Referable diabetic retinopathy detection in Kaggle-test and e-ophtha-test using `net B'. Kaggle-test-10000 denotes a subset of 10,000 images from Kaggle-test.}
  \label{fig:kaggle_imageLevel}
\end{figure}

Random forest parameters, the number of trees $n_T$ and the maximum depth of each tree $d_T$, were optimized by 5-fold cross-validation in Kaggle-train and e-ophtha-train: $n_T = 500$ in Kaggle DR dataset and $300$ in e-ophtha, $d_T = 25$ in both datasets. ROC curves in Kaggle-test and e-ophtha-test are reported in Fig. \ref{fig:kaggle_imageLevel} for `net B'. Areas under the ROC curves for all three ConvNets are summarized in Table \ref{tab:kaggle_imageLevel}. It is important to notice that using such an ensemble of ConvNets does not necessarily turn the solution into a black box. Indeed, the heatmaps associated with the above ConvNets all have the same size. So, to support decision for a given patient, the heatmaps associated with all the networks involved in the decision process can be blended (one blended heatmap per eye). Furthermore, each heatmap can be weighted by the importance of the associated ConvNet in the decision process.
\begin{table}
  \caption{Referable diabetic retinopathy detection in Kaggle-test and e-ophtha-test --- area under the ROC curve.}
  \begin{center}
  {
    \footnotesize
    \begin{tabular}{c|c||c|c}
    ensemble                     & ConvNet & Kaggle-test & e-ophtha-test \\
    \hline
    \multirow{3}{*}{no}   & net A     & 0.940           & $\emptyset$ \\
                                        & net B     & 0.944           & $\emptyset$ \\
                                        & AlexNet & 0.900           & $\emptyset$ \\
    \hline
    \multirow{4}{*}{yes} & net A     & 0.952           & 0.948 \\
                                       & net B     & 0.954            & 0.949 \\
                                       & AlexNet & 0.908            & 0.907 \\
                                       & all          & 0.955            & 0.949 \\
    \end{tabular}
  }
  \end{center}
  \label{tab:kaggle_imageLevel}
\end{table}

\begin{figure*}
  \begin{center}
    \begin{tabular}{cc}
      \subfloat[original image]{
        \includegraphics[width=0.34\textwidth]{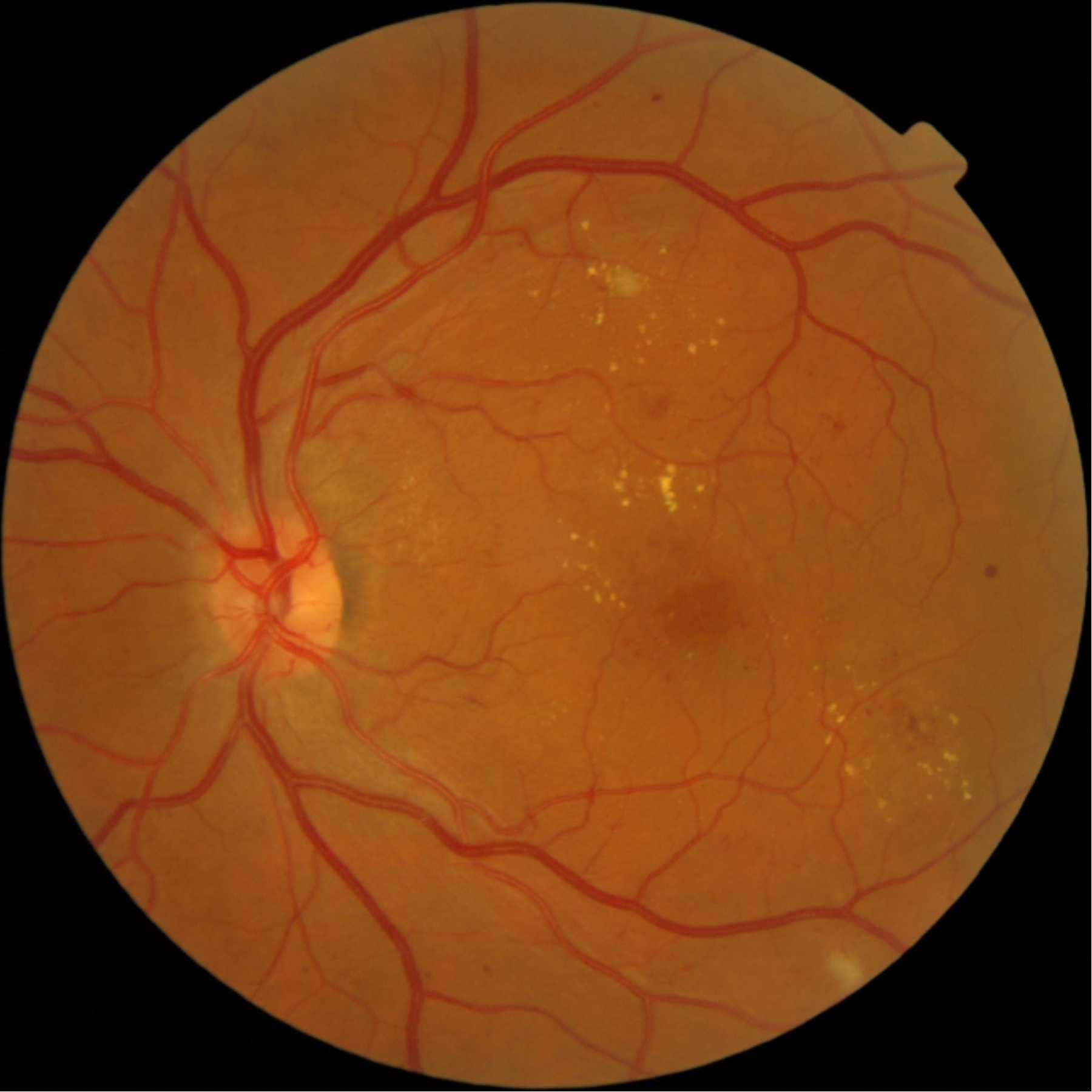}
      } &
      \subfloat[preprocessed image]{
        \includegraphics[width=0.34\textwidth]{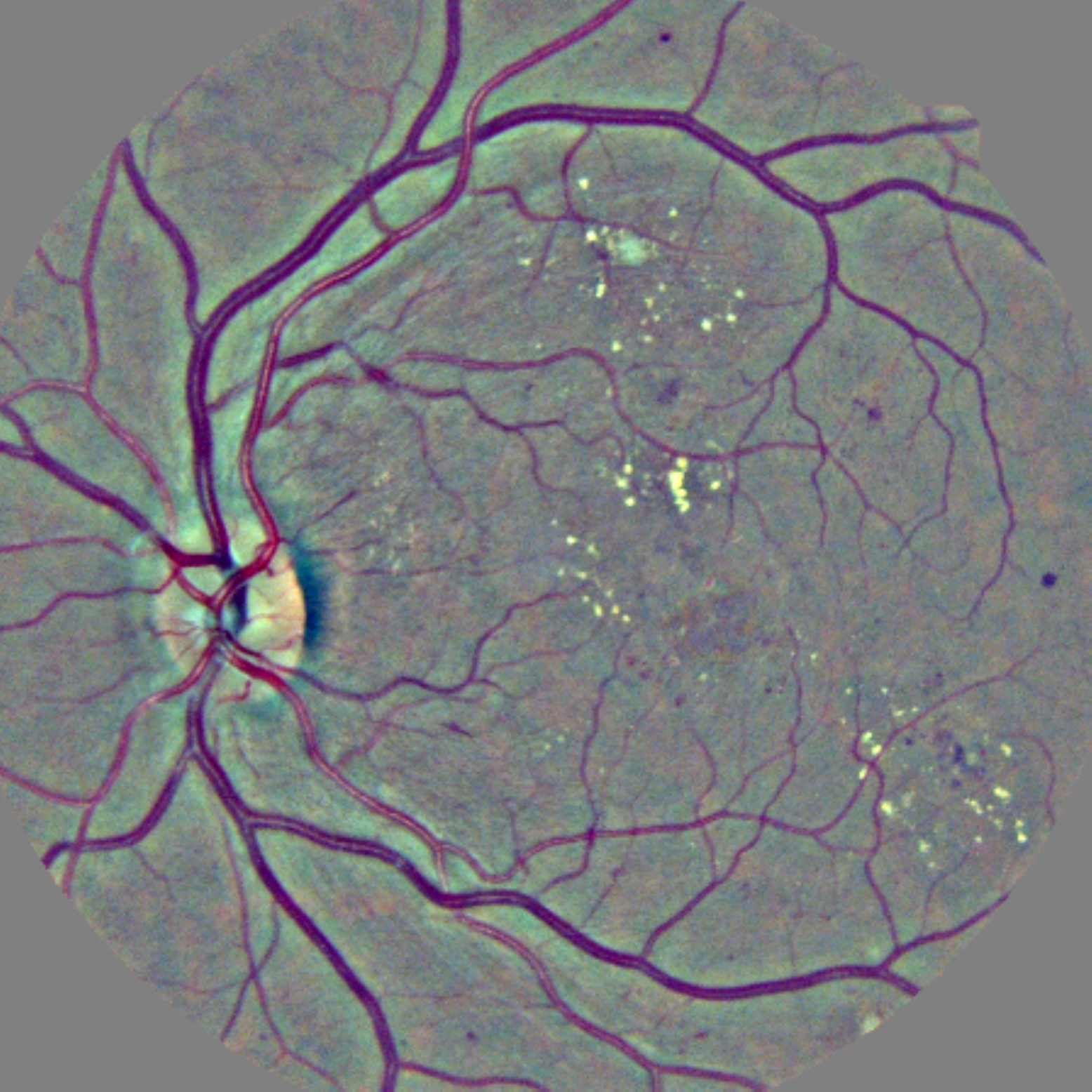}
      } \\
      \subfloat[$\pi_{HE}$ --- mostly bright lesions]{
        \includegraphics[width=0.34\textwidth]{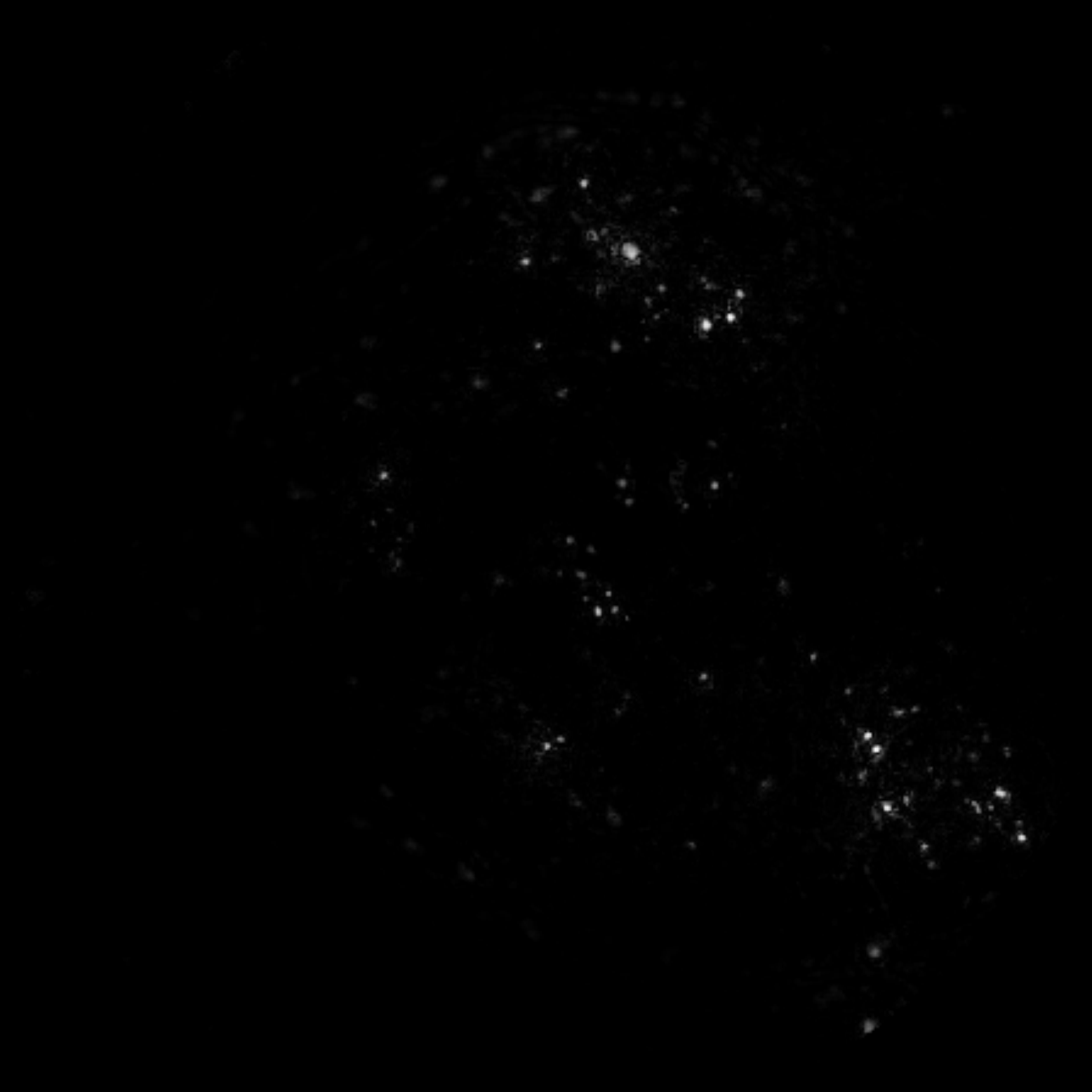}
      } &
      \subfloat[$\pi_{RDR}$ --- mostly red lesions]{
        \includegraphics[width=0.34\textwidth]{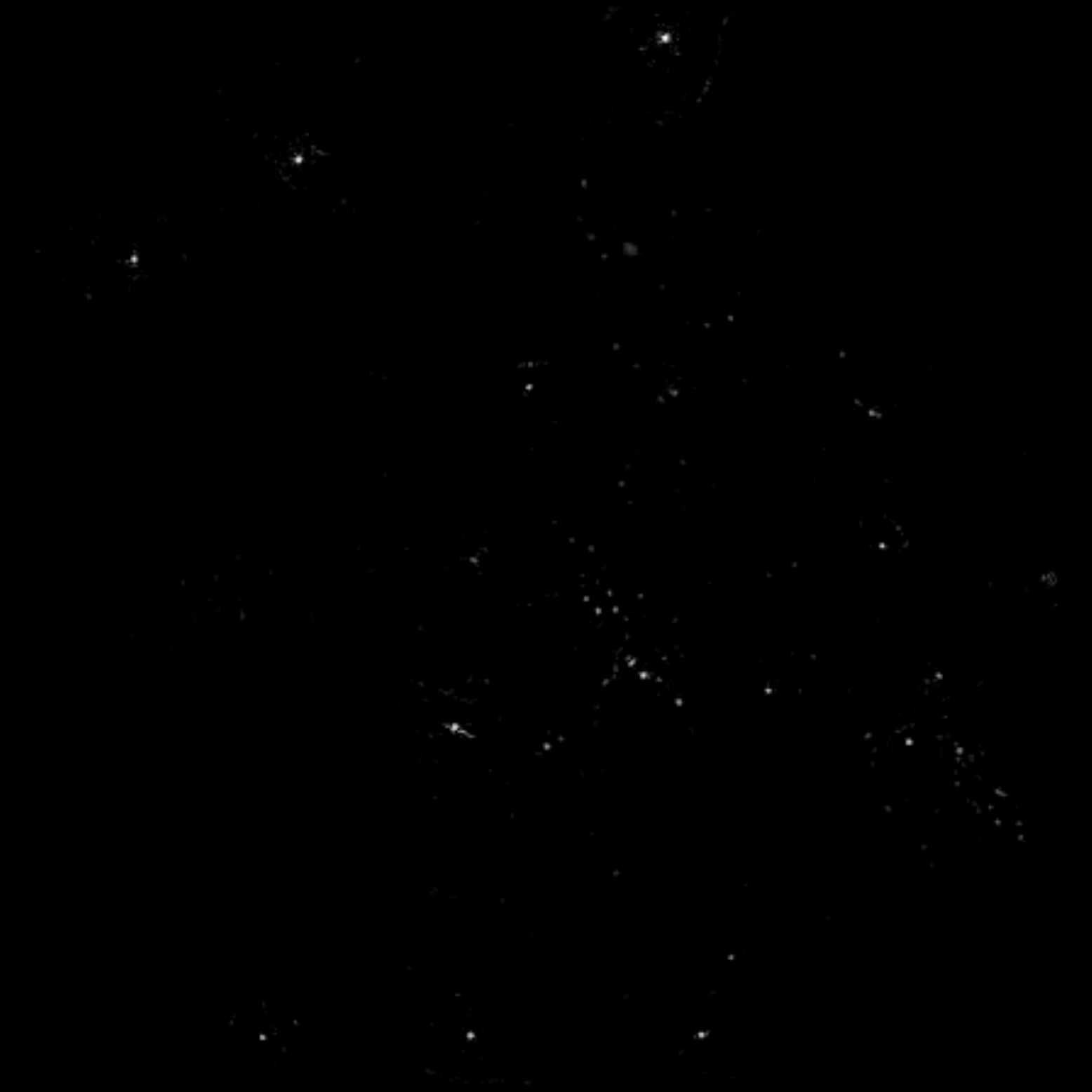}
      } \\
      \subfloat[$\bar{\pi}$ --- all lesions]{
        \includegraphics[width=0.34\textwidth]{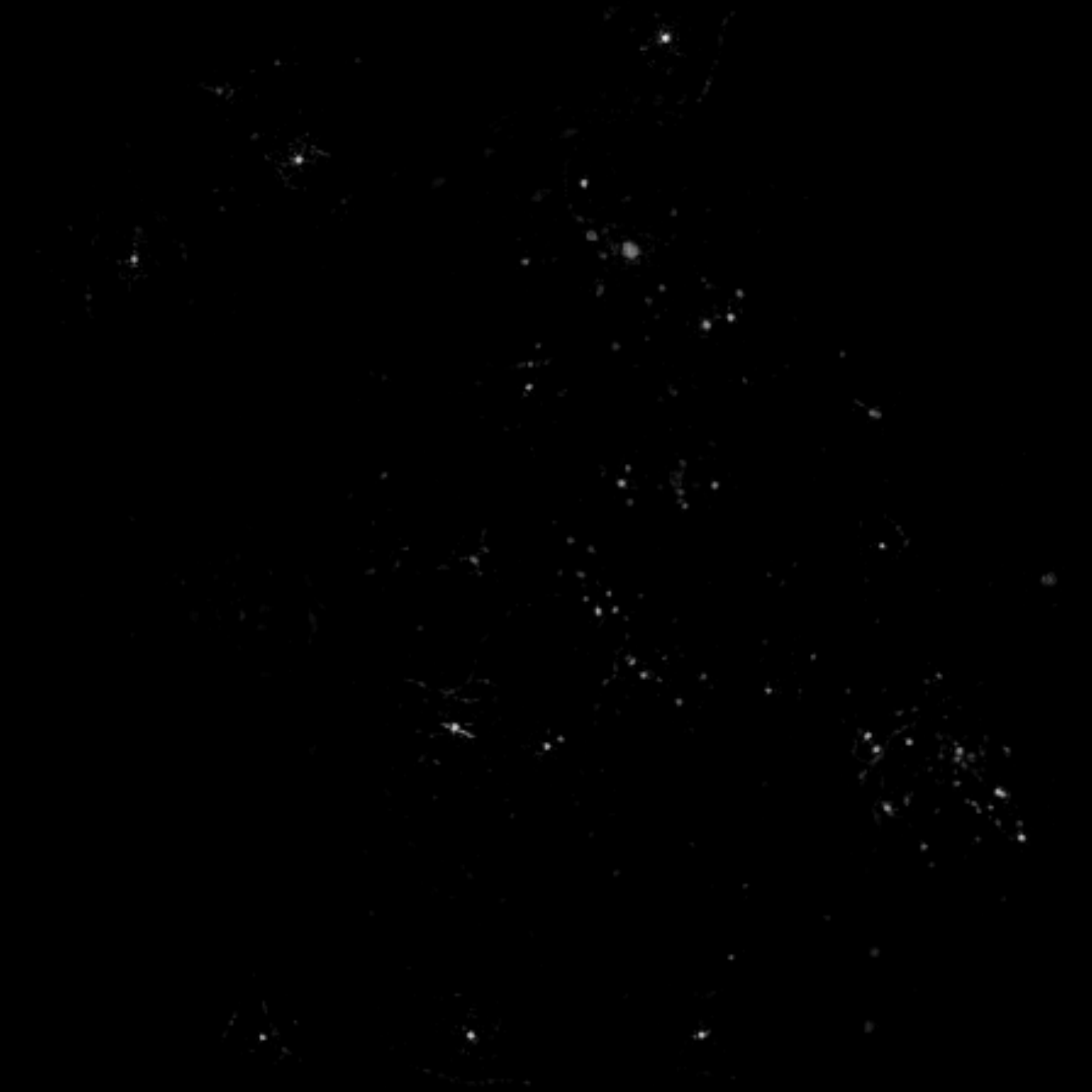}
      } &
      \subfloat[$\pi_0$ --- all lesions, but also blood vessels]{
        \includegraphics[width=0.34\textwidth]{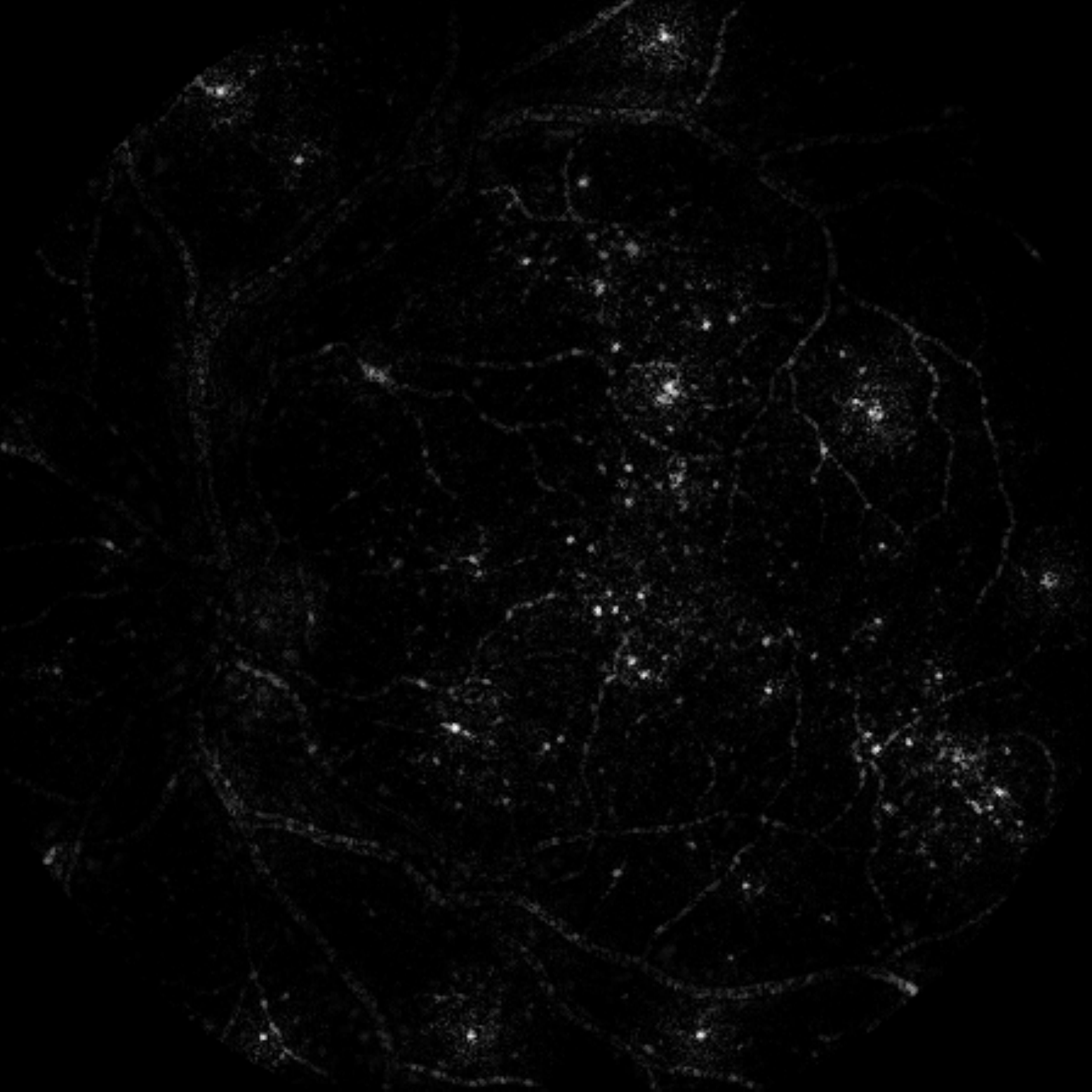}
      } \\
    \end{tabular}
  \end{center}
  \caption{Examples of heatmaps from a fundus image of typical quality (image 20051020\_57622\_0100\_PP.png from Messidor). $\pi_0$ was obtained without sparsity enhancement. $\pi_{HE}$ and $\pi_{RDR}$ were obtained with sparsity enhancement, using networks $\mathcal{N}_{HE}$ and $\mathcal{N}_{RDR}$, respectively. $\bar{\pi}$ was obtained by averaging six $\pi$ maps (including $\pi_{HE}$ and $\pi_{RDR}$), where each $\pi$ map is weighted according to the network's importance in the random forest.}
  \label{fig:example1}
\end{figure*}

\begin{figure*}
  \begin{center}
    \begin{tabular}{cc}
      \subfloat[original image]{
        \includegraphics[width=0.34\textwidth]{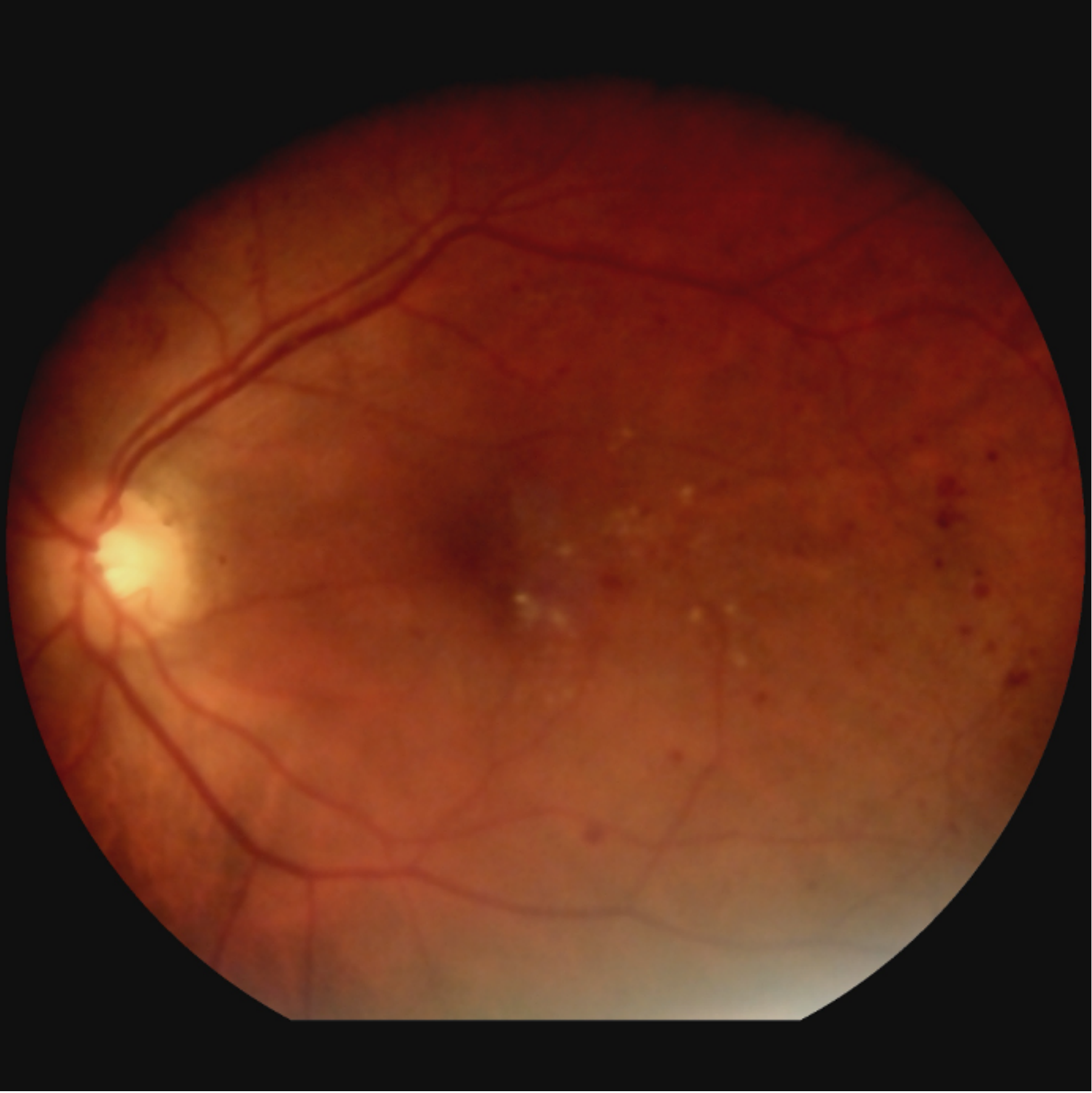}
      } &
      \subfloat[preprocessed image]{
        \includegraphics[width=0.34\textwidth]{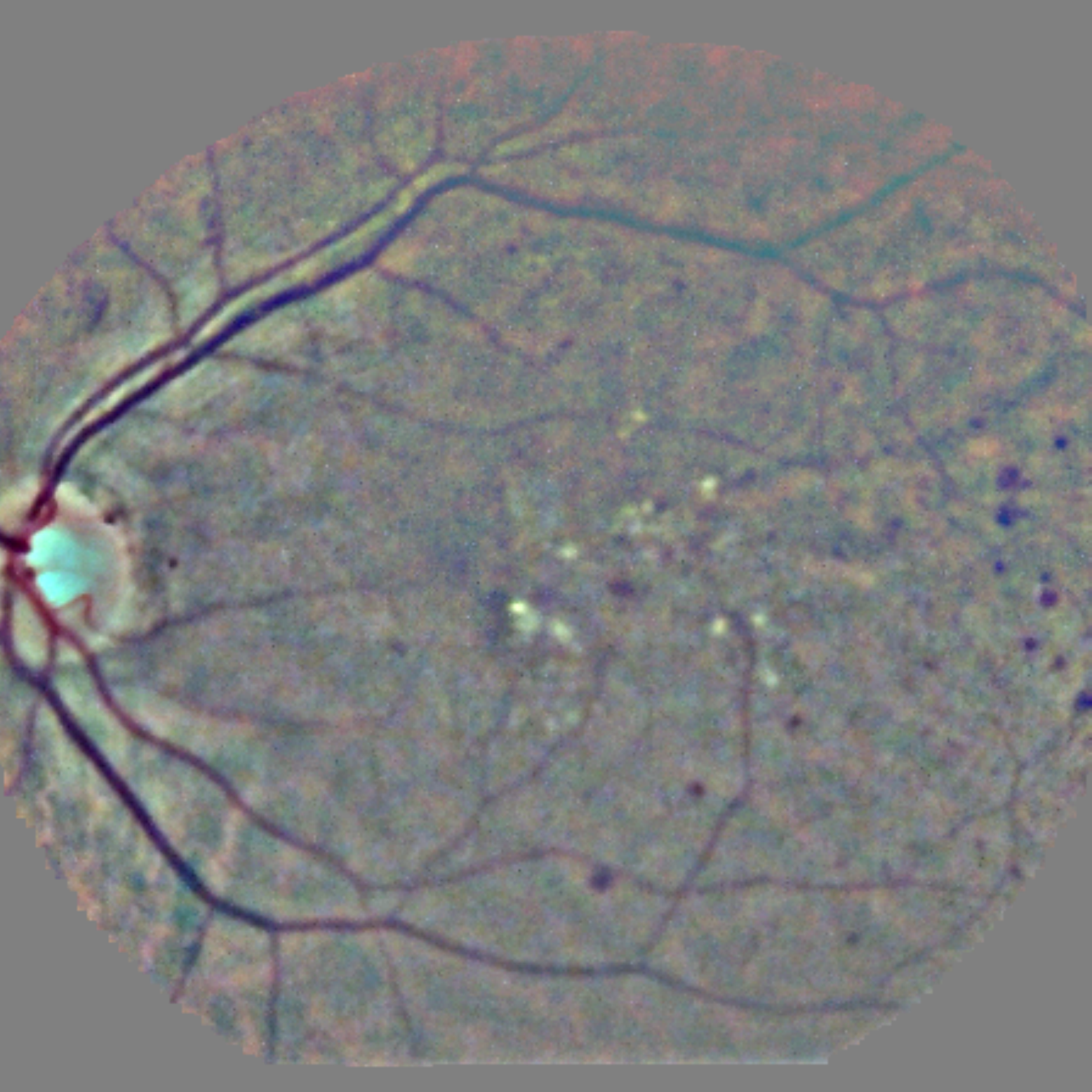}
      } \\
      \subfloat[$\pi_{HE}$ --- mostly bright lesions]{
        \includegraphics[width=0.34\textwidth]{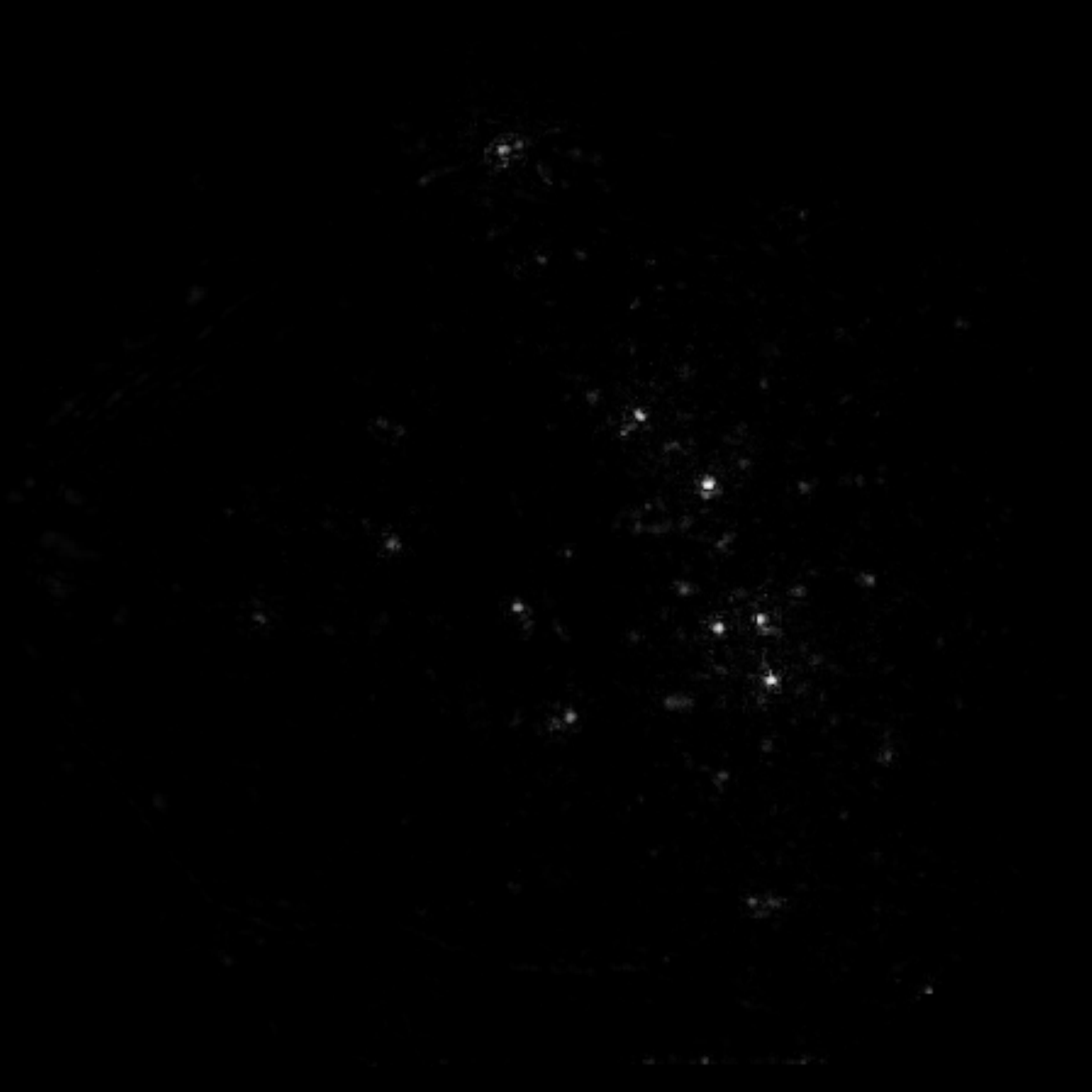}
      } &
      \subfloat[$\pi_{RDR}$ --- mostly red lesions]{
        \includegraphics[width=0.34\textwidth]{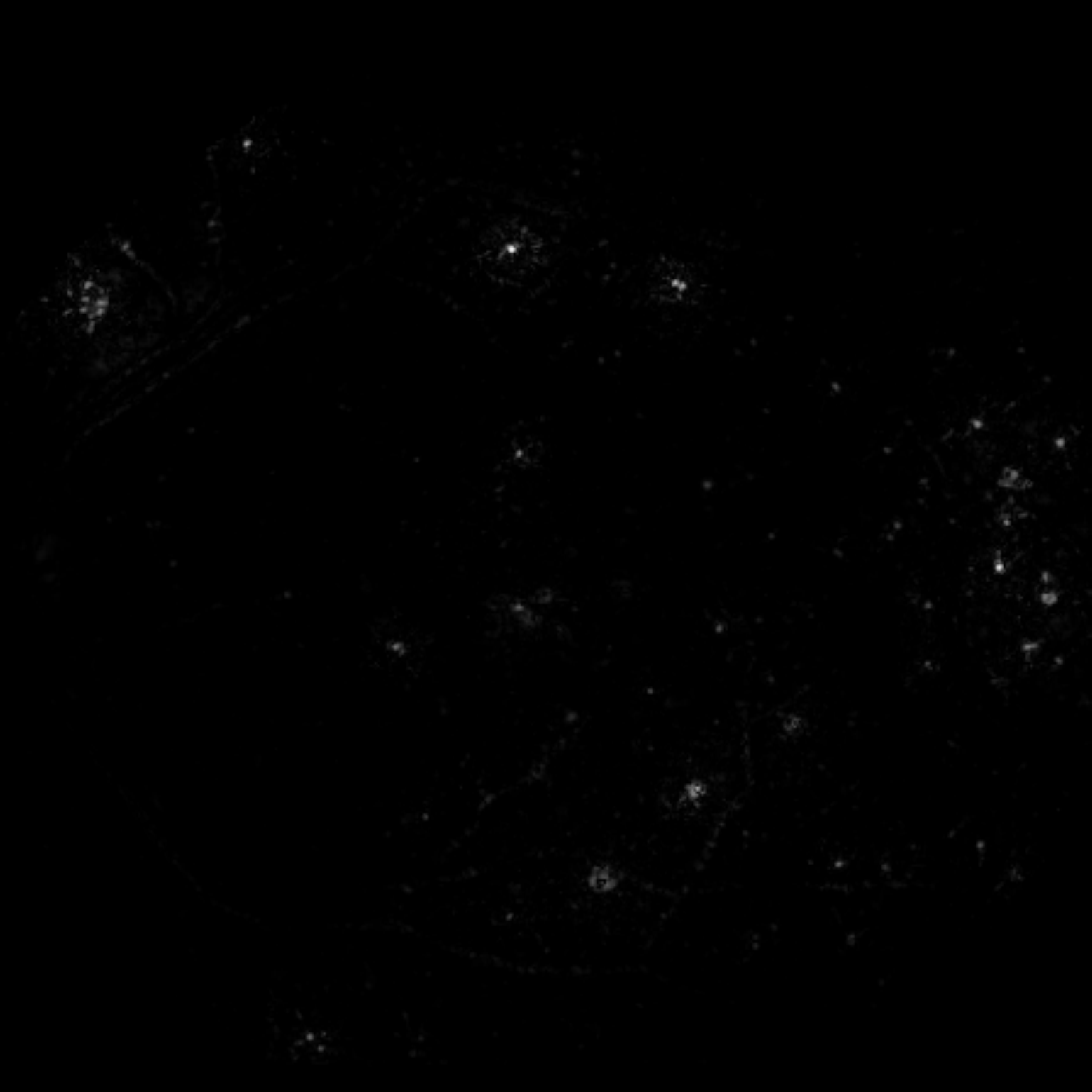}
      } \\
      \subfloat[$\bar{\pi}$ --- all lesions]{
        \includegraphics[width=0.34\textwidth]{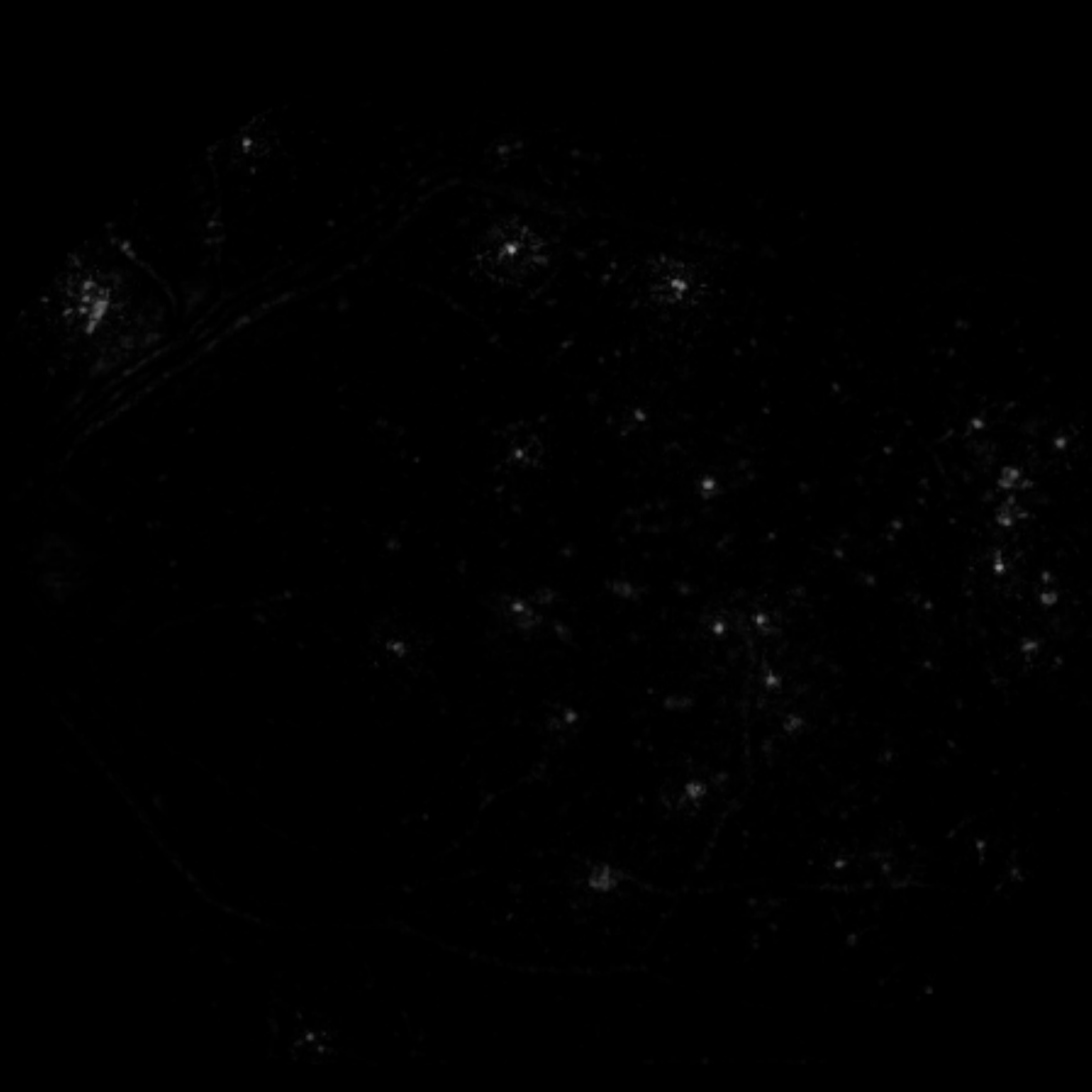}
      } &
      \subfloat[$\pi_0$ --- all lesions, but also blood vessels]{
        \includegraphics[width=0.34\textwidth]{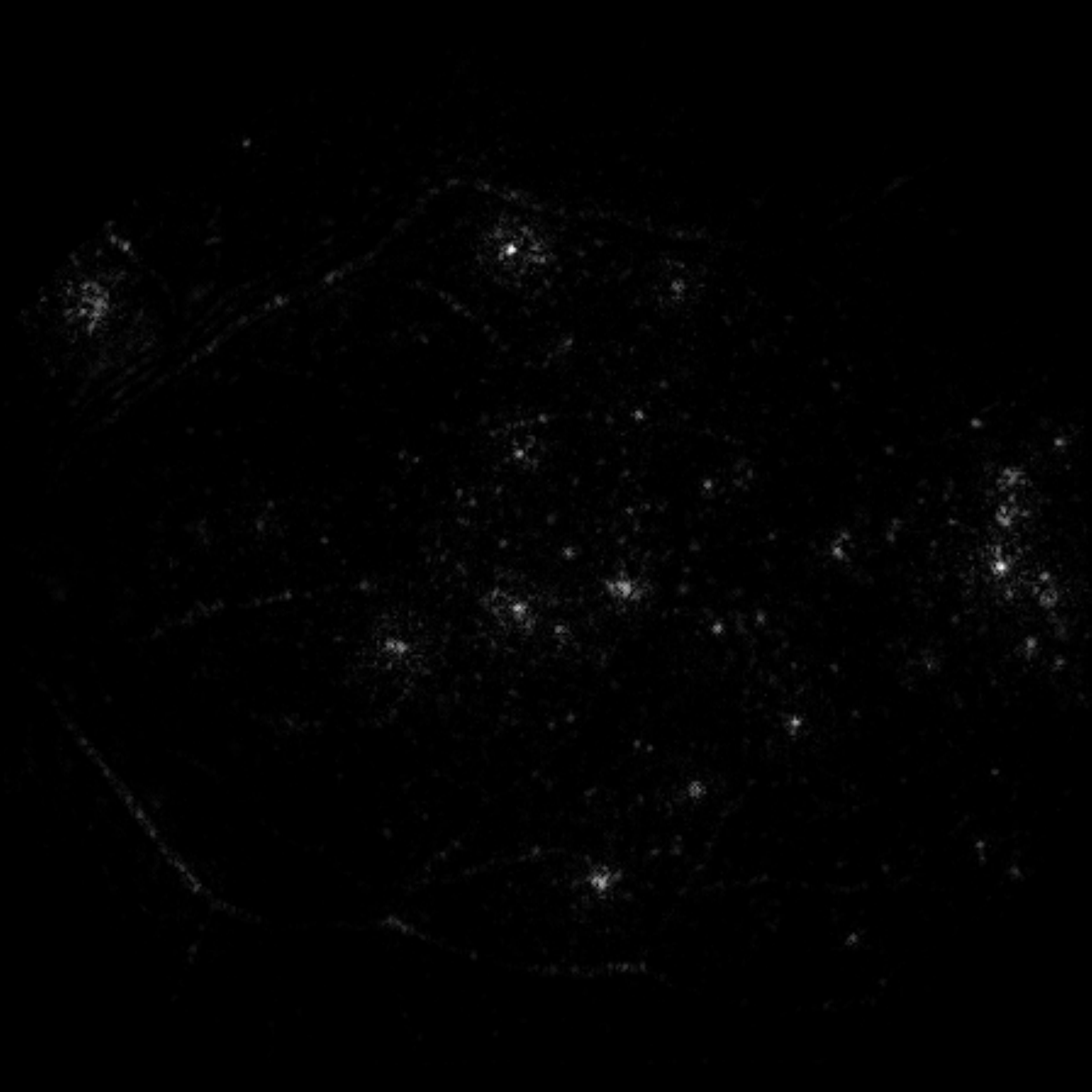}
      } \\
    \end{tabular}
  \end{center}
  \caption{Examples of heatmaps from a fundus image acquired with a low-cost handheld retinograph: the Horus DEC 200 (MiiS, Hsinchu, Taiwan). This image was acquired at Brest University Hospital in 2015 \citep{quellec_suitability_2016}.}
  \label{fig:example2}
\end{figure*}

The performance of the proposed pixel-level detector, using `net B', is illustrated in Fig. \ref{fig:example1} and \ref{fig:example2} on two images from independent datasets. The first image comes from the publicly-available Messidor dataset\footnote{\url{http://www.adcis.net/en/Download-Third-Party/Messidor.html}}. The second image comes from a private dataset acquired with a low-cost handheld retinograph  \citep{quellec_suitability_2016}.

\section{Discussion and Conclusions}
\label{sec:DiscussionConclusions}

A solution was proposed in this paper for the automatic detection of referable diabetic retinopathy (DR) and for the automatic detection of lesions related to DR. Unlike competing solutions, this lesion detector is trained using image-level labels only for supervision. The solution relies on ConvNets trained to detect referable DR at the image level. Using a modified sensitivity analysis, the pixels playing a role in the image-level predictions are detected: a heatmap the size of the image is obtained. In order to improve the quality of the heatmaps (attenuate artifacts), we proposed to enhance the sparsity of the heatmaps while training the ConvNets. Because those heatmaps depend on backpropagated quantities, the network parameters cannot be optimized using the usual backpropagation method, so a different ConvNet training method was proposed.

Three ConvNets were trained to detect referable DR in the Kaggle-train dataset, using the proposed heatmap optimization procedure. Then, we evaluated how well those ConvNets could detect lesions in the DiaretDB1 dataset, without retraining them. For lesion detection at the image level, they outperformed previous algorithms, which were explicitly trained to detect the target lesions, with pixel-level supervision (see Fig. \ref{fig:diaretdb1_roc}). This superiority was observed for all lesions or groups of lesions, with the exception of `red lesions'. Experiments were also performed at the lesion level: for all lesion types, the proposed algorithm was found to outperform recent heatmap generation algorithms (see Table \ref{tab:diaretdb1_froc}). As illustrated in two examples (see Fig. \ref{fig:example1} and \ref{fig:example2}), the produced heatmaps are of very good quality. In particular, the false alarms detected on the vessels, in the vicinity of true lesions in the unoptimized heatmaps ($\pi_0$ maps), are strongly reduced with sparsity maximization ($\pi_{HE}$, $\pi_{RDR}$, $\bar{\pi}$). These experiments validate the relevance of image-level supervision for lesion detectors, but stress the need to optimize the heatmaps, as proposed in this paper. Note that detection performance is not affected much by image quality: very good detections are produced in the blurry image obtained with a low-cost, handheld retinograph (see Fig. \ref{fig:example2}). This is a very important feature, which opens the way to automated mobile screening. However, it can be observed that the `AlexNet' architecture, which achieves moderate DR detection results, also achieves poor detection results at the lesion level, even after heatmap optimization (see Table \ref{tab:diaretdb1_froc}): to ensure good detection performance at the lesion level, the proposed optimization framework should be applied to ConvNet architectures that achieve good image-level performance.

Besides analyzing the pixel-level performance of the final ConvNets, we also analyzed the pixel-level performance while the ConvNets are being trained (see Fig. \ref{fig:validationPerformance} for the `net B' ConvNet). It turned out to be quite instructive. By analyzing performance at the image level alone (the area under the ROC curve in the Kaggle validation set), all we can see is that 1) performance quickly reaches a satisfactorily level ($A_z \simeq 0.85$), then 2) slowly increases for many iterations, 3) experiences a second leap to reach a very high level ($A_z \simeq 0.93$) and finally 4) reaches its optimal state ($A_z \simeq 0.95$) very slowly. By analyzing the heatmaps, we understand that the ConvNet very quickly learns to detect exudates and cotton-wool spots (or `soft exudates'). The second performance leap is observed when the ConvNet learns to detect hemorrhages. The final fine-tuning stage correlates with the progressive discovery of microaneurysms (or `small red dots') by the ConvNet. Interestingly, lesions were discovered in the same order regardless of the ConvNet structure (`net A', `net B' or AlexNet). The order in which lesions are discovered by ConvNets makes sense: the most obvious lesions (the largest and the most contrasted) are detected first and then the ConvNet discovers that more subtle lesions (which are more difficult to capture) are in fact more useful to make a diagnosis. By the way, for `net A' and `net B', we observe that the detection performance of bright lesions (exudates and cotton-wool spots) decreases when red lesions (hemorrhages and microaneurysms) are discovered: red lesions are indeed usually enough for detecting referable DR. This behavior is not observed for AlexNet: the reason probably is that red lesions are not detected well enough (see Table \ref{tab:diaretdb1_froc}), due to the low resolution of input images (224 $\times$ 224 pixels). The main difference between the two plots of Fig. \ref{fig:validationPerformance} (with or without sparsity maximization) is observed in the latest stages of training. As the artifacts are removed by enhancing the sparsity of the heatmaps, the detection performance at the pixel level increases for every lesion type. We hypothesized that maximizing the sparsity of the heatmaps would also speed up the training process, by reducing the search space. However, we did not observe such a behavior.

Performance at the image level is also very high, which was to be expected since we used efficient ConvNets from a recent machine learning competition as starting points: a performance of $A_z = 0.9542$ was achieved in Kaggle-test by the proposed framework using `net B' (95 \% confidence interval with \citet{delong_comparing_1988}'s method: $[0.9515, 0.9569]$). In particular, the proposed solution outperforms the system by \citet{colas_deep_2016} ($A_z = 0.946$). This good performance derives in part from the above observations at the pixel level, which explains that the proposed solution also outperforms our main baseline solution, namely o\_O ($A_z = 0.9512$). The performance of the ensemble was also very high in e-ophtha-test: $A_z = 0.9490$ (95 \% confidence interval: $[0.9459, 0.9521]$), as opposed to $A_z = 0.8440$ for our previous solution, based in part on multiple-instance learning \citep{quellec_automatic_2016}. The proposed ensemble strategy was extended to multiple network structures, but it did not increase performance significantly (see Table \ref{tab:kaggle_imageLevel}): because it increases complexity for a very limited benefit, we do not recommend it. It should be noted that \citet{gulshan_development_2016} recently reported higher performance (up to $A_z = 0.991$) in smaller datasets (less than 5,000 patients) with a much better ground truth (seven or eight grades per image, as opposed to one in this study); that system does not provide pixel-level information. \citet{abramoff_improved_2016} also reported higher performance ($A_z = 0.980$) in a smaller dataset (874 patients) with a better ground truth (three grades per image) for a system relying on pixel-level information for training.

In this study, we focused on detecting referable DR for several reasons. First, this is the most useful classification in the context of DR screening: it determines whether or not the patient needs to be seen in person by an ophthalmologist. Second, the manual segmentations used to evaluate performance at the pixel level do not contain the most advanced signs of DR, such as neovascularizations, so additional manual segmentations would be needed for a relevant evaluation. Third, it allows comparison with the state of the art \citep{colas_deep_2016}. However, unlike most multiple-instance learning algorithms, the proposed solution is not restricted to binary decision problems. The proposed solution is also general in the sense that it could be applied advantageously to all problems governed by a very limited number of relevant pixels. This class of problems was traditionally solved by multiple-instance Learning algorithms, but the proposed solution allows these problems to be solved by ConvNets as well, with enhanced performance. Finally, because the proposed framework is trained to detect relevant patterns in images using image-level labels only for supervision, it can be regarded as a general image mining tool, hence the paper's title. In particular, this framework has the potential to discover new biomarkers in images, which we will try to demonstrate in future works.

The solution presented in this paper for learning to detect referable DR and DR lesions does not require neither expert knowledge nor expert segmentations: it only requires referral decisions stored in examination records. However, expert segmentations (from DiaretDB1) helped us fine-tune the system and improve its performance further. Also, \citet{gulshan_development_2016} showed that increasing the number of grades per training image, by soliciting multiple experts, significantly improved the performance of their deep learning system. In other words, deep learning solutions will always benefit from clinicians for training, and also of course for assessing their predictions.

In conclusion, a general framework was proposed for solving multiple-instance problems with ConvNets and the relevance of this framework was demonstrated in the context of diabetic retinopathy screening.

\section{Acknowledgements}

This work was supported in part by a grant from the French \textit{Fond Unique Interminist\'eriel} (FUI-19 RetinOpTIC). The authors would also like to thank the organizers and competitors of the Kaggle Diabetic Retinopathy competition for providing very useful data and ideas.

\bibliographystyle{elsarticle-harv}
\bibliography{deepImageMining}

\begin{thebibliography}{53}
\expandafter\ifx\csname natexlab\endcsname\relax\def\natexlab#1{#1}\fi
\expandafter\ifx\csname url\endcsname\relax
  \def\url#1{\texttt{#1}}\fi
\expandafter\ifx\csname urlprefix\endcsname\relax\def\urlprefix{URL }\fi

\bibitem[{Abr{\`a}moff et~al.(2010)Abr{\`a}moff, Garvin, and
  Sonka}]{abramoff_retinal_2010}
Abr{\`a}moff, M.~D., Garvin, M.~K., Sonka, M., 2010. Retinal imaging and image
  analysis. IEEE Rev Biomed Eng 3, 169--208.

\bibitem[{Abr{\`a}moff et~al.(2016)Abr{\`a}moff, Lou, Erginay, Clarida, Amelon,
  Folk, and Niemeijer}]{abramoff_improved_2016}
Abr{\`a}moff, M.~D., Lou, Y., Erginay, A., Clarida, W., Amelon, R., Folk,
  J.~C., Niemeijer, M., Oct. 2016. Improved automated detection of diabetic
  retinopathy on a publicly available dataset through integration of deep
  learning. Invest Ophthalmol Vis Sci 57~(13), 5200--5206.

\bibitem[{Arunkumar and Karthigaikumar(2017)}]{arunkumar_multi-retinal_2015}
Arunkumar, R., Karthigaikumar, P., Feb. 2017. Multi-retinal disease
  classification by reduced deep learning features. Neural Comput Appl 28~(2),
  329--334.

\bibitem[{Bach et~al.(2015)Bach, Binder, Montavon, Klauschen, M{\"u}ller, and
  Samek}]{bach_pixel-wise_2015}
Bach, S., Binder, A., Montavon, G., Klauschen, F., M{\"u}ller, K.-R., Samek,
  W., Jul. 2015. On pixel-wise explanations for non-linear classifier decisions
  by layer-wise relevance propagation. PLoS One 10~(7).

\bibitem[{Barriga et~al.(2014)Barriga, McGrew, VanNess, Zamora, Nemeth, Bauman,
  and Soliz}]{barriga_assessing_2014}
Barriga, E.~S., McGrew, E., VanNess, R., Zamora, G., Nemeth, S.~C., Bauman, W.,
  Soliz, P., Apr. 2014. Assessing agreement between dilated indirect
  stereoscopic exam and digital non-mydriatic retinal photography for the
  evaluation of diabetic retinopathy. In: Proc {ARVO}. Vol.~55. Orlando, FL,
  USA, pp. 5335--5335.

\bibitem[{Bharali et~al.(2015)Bharali, Medhi, and
  Nirmala}]{bharali_detection_2015}
Bharali, P., Medhi, J., Nirmala, S., 2015. Detection of hemorrhages in diabetic
  retinopathy analysis using color fundus images. In: Proc {IEEE} {ReTIS}.
  Kolkata, India, pp. 237--242.

\bibitem[{Burlina et~al.(2016)Burlina, Freund, Joshi, Wolfson, and
  Bressler}]{burlina_detection_2016}
Burlina, P., Freund, D., Joshi, N., Wolfson, Y., Bressler, N., Apr. 2016.
  Detection of age-related macular degeneration via deep learning. In: Proc
  {ISBI}. Prague, Czech Republic, pp. 184--188.

\bibitem[{Chen et~al.(2015{\natexlab{a}})Chen, Xu, Wong, Wong, and
  Liu}]{chen_glaucoma_2015}
Chen, X., Xu, Y., Wong, D., Wong, T., Liu, J., Aug. 2015{\natexlab{a}}.
  Glaucoma detection based on deep convolutional neural network. In: Proc
  {IEEE} {EMBC}. Milan, Italy, pp. 715--718.

\bibitem[{Chen et~al.(2015{\natexlab{b}})Chen, Xu, Yan, Wong, Wong, and
  Liu}]{chen_automatic_2015}
Chen, X., Xu, Y., Yan, S., Wong, D., Wong, T., Liu, J., Oct.
  2015{\natexlab{b}}. Automatic feature learning for glaucoma detection based
  on deep learning. In: Proc {MICCAI}. Munich, Germany, pp. 669--677.

\bibitem[{Colas et~al.(2016)Colas, Besse, Orgogozo, Schmauch, Meric, and
  Besse}]{colas_deep_2016}
Colas, E., Besse, A., Orgogozo, A., Schmauch, B., Meric, N., Besse, E., Oct.
  2016. Deep learning approach for diabetic retinopathy screening. In: Froc
  {EVER}. Nice, France.

\bibitem[{Cuadros and Bresnick(2009)}]{cuadros_eyepacs:_2009}
Cuadros, J., Bresnick, G., May 2009. {EyePACS}: an adaptable telemedicine
  system for diabetic retinopathy screening. J Diabetes Sci Technol 3~(3),
  509--516.

\bibitem[{Dai et~al.(2016)Dai, Bu, Wang, and Wu}]{dai_fundus_2016}
Dai, B., Bu, W., Wang, K., Wu, X., 2016. Fundus lesion detection based on
  visual attention model. Commun Comput Inform Sci 623, 384--394.

\bibitem[{DeLong et~al.(1988)DeLong, DeLong, and
  Clarke-Pearson}]{delong_comparing_1988}
DeLong, E.~R., DeLong, D.~M., Clarke-Pearson, D.~L., Sep. 1988. Comparing the
  areas under two or more correlated receiver operating characteristic curves:
  a nonparametric approach. Biometrics 44~(3), 837--845.

\bibitem[{Erginay et~al.(2008)Erginay, Chabouis, Viens-Bitker, Robert,
  Lecleire-Collet, and Massin}]{Erginay2008}
Erginay, A., Chabouis, A., Viens-Bitker, C., Robert, N., Lecleire-Collet, A.,
  Massin, P., Jun 2008. {OPHDIAT}: quality-assurance programme plan and
  performance of the network. Diabetes Metab 34~(3), 235--42.

\bibitem[{Franklin and Rajan(2014)}]{franklin_diagnosis_2014}
Franklin, S., Rajan, S., 2014. Diagnosis of diabetic retinopathy by employing
  image processing technique to detect exudates in retinal images. IET Image
  Process 8~(10), 601--609.

\bibitem[{Girshick et~al.(2014)Girshick, Donahue, Darrell, and
  Malik}]{girshick_rich_2014}
Girshick, R., Donahue, J., Darrell, T., Malik, J., 2014. Rich feature
  hierarchies for accurate object detection and semantic segmentation. In: Proc
  {IEEE} {CVPR}. Washington, DC, USA, pp. 580--587.

\bibitem[{Goodfellow et~al.(2013)Goodfellow, Warde-Farley, Mirza, Courville,
  and Bengio}]{goodfellow_maxout_2013}
Goodfellow, I.~J., Warde-Farley, D., Mirza, M., Courville, A., Bengio, Y., Jun.
  2013. Maxout networks. In: Proc {ICML}. Vol.~28. Atlanta, GA, USA, pp.
  1319--1327.

\bibitem[{Graham(2014)}]{graham_fractional_2014}
Graham, B., Dec. 2014. Fractional max-pooling. Tech. Rep. arXiv:1412.6071 [cs].

\bibitem[{Gulshan et~al.(2016)Gulshan, Peng, Coram, Stumpe, Wu, Narayanaswamy,
  Venugopalan, Widner, Madams, Cuadros, Kim, Raman, Nelson, Mega, and
  Webster}]{gulshan_development_2016}
Gulshan, V., Peng, L., Coram, M., Stumpe, M.~C., Wu, D., Narayanaswamy, A.,
  Venugopalan, S., Widner, K., Madams, T., Cuadros, J., Kim, R., Raman, R.,
  Nelson, P.~C., Mega, J.~L., Webster, D.~R., Nov. 2016. Development and
  validation of a deep learning algorithm for detection of diabetic retinopathy
  in retinal fundus photographs. JAMA.

\bibitem[{Haloi(2015)}]{haloi_improved_2015}
Haloi, M., May 2015. Improved microaneurysm detection using deep neural
  networks. Tech. Rep. arXiv:1505.04424 [cs].

\bibitem[{He et~al.(2015)He, Zhang, Ren, and Sun}]{he_delving_2015}
He, K., Zhang, X., Ren, S., Sun, J., Feb. 2015. Delving deep into rectifiers:
  surpassing human-level performance on {ImageNet} classification. Tech. Rep.
  arXiv:1502.01852 [cs.CV].

\bibitem[{Hinton et~al.(2012)Hinton, Srivastava, Krizhevsky, Sutskever, and
  Salakhutdinov}]{hinton_improving_2012}
Hinton, G.~E., Srivastava, N., Krizhevsky, A., Sutskever, I., Salakhutdinov,
  R.~R., Jul. 2012. Improving neural networks by preventing co-adaptation of
  feature detectors. Tech. Rep. arXiv:1207.0580 [cs].

\bibitem[{Kauppi et~al.(2007)Kauppi, Kalesnykiene, Kamarainen, Lensu, Sorri,
  Raninen, Voutilainen, Pietil{\"a}, K{\"a}lvi{\"a}inen, and
  Uusitalo}]{kauppi_diaretdb1_2007}
Kauppi, T., Kalesnykiene, V., Kamarainen, J.-K., Lensu, L., Sorri, I., Raninen,
  A., Voutilainen, R., Pietil{\"a}, J., K{\"a}lvi{\"a}inen, H., Uusitalo, H.,
  2007. The {DIARETDB}1 diabetic retinopathy database and evaluation protocol.
  In: Proc {BMVC}. Warwik, UK.

\bibitem[{Kingma and Ba(2015)}]{kingma_adam:_2015}
Kingma, D., Ba, J., May 2015. Adam: a method for stochastic optimization. In:
  Proc {ICLR}. San Diego, CA, USA.

\bibitem[{Krizhevsky et~al.(2012)Krizhevsky, Sutskever, and
  Hinton}]{krizhevsky_imagenet_2012}
Krizhevsky, A., Sutskever, I., Hinton, G.~E., 2012. {ImageNet} classification
  with deep convolutional neural networks. In: Proc {Adv} {Neural} {Inform}
  {Process} {Syst}. Vol.~25. Granada, Spain, pp. 1097--1105.

\bibitem[{Kumar et~al.(2014)Kumar, Kumar, Sathar, and
  Sahasranamam}]{kumar_automatic_2014}
Kumar, P., Kumar, R., Sathar, A., Sahasranamam, V., 2014. Automatic detection
  of red lesions in digital color retinal images. In: Proc {IC}3I. Mysore,
  India, pp. 1148--1153.

\bibitem[{LeCun et~al.(2015)LeCun, Bengio, and Hinton}]{lecun_deep_2015}
LeCun, Y., Bengio, Y., Hinton, G., May 2015. Deep learning. Nature 521~(7553),
  436--444.

\bibitem[{Li et~al.(2016)Li, Feng, Xie, Liang, Zhang, and
  Wang}]{li_cross-modality_2016}
Li, Q., Feng, B., Xie, L., Liang, P., Zhang, H., Wang, T., Jan. 2016. A
  cross-modality learning approach for vessel segmentation in retinal images.
  IEEE Trans Med Imaging 35~(1), 109--118.

\bibitem[{Lim et~al.(2015)Lim, Cheng, Hsu, and Lee}]{lim_integrated_2015}
Lim, G., Cheng, Y., Hsu, W., Lee, M., Nov. 2015. Integrated optic disc and cup
  segmentation with deep learning. In: Proc {ICTAI}. Vietri sul Mare, Italy,
  pp. 162--169.

\bibitem[{Maji et~al.(2015)Maji, Santara, Ghosh, Sheet, and
  Mitra}]{maji_deep_2015}
Maji, D., Santara, A., Ghosh, S., Sheet, D., Mitra, P., Aug. 2015. Deep neural
  network and random forest hybrid architecture for learning to detect retinal
  vessels in fundus images. In: Proc {IEEE} {EMBC}. Milan, Italy, pp.
  3029--3032.

\bibitem[{Maji et~al.(2016)Maji, Santara, Mitra, and
  Sheet}]{maji_ensemble_2016}
Maji, D., Santara, A., Mitra, P., Sheet, D., Mar. 2016. Ensemble of deep
  convolutional neural networks for learning to detect retinal vessels in
  fundus images. Tech. Rep. arXiv:1603.04833 [cs, stat].

\bibitem[{Mane et~al.(2015)Mane, Kawadiwale, and Jadhav}]{mane_detection_2015}
Mane, V., Kawadiwale, R., Jadhav, D., 2015. Detection of red lesions in
  diabetic retinopathy affected fundus images. In: Proc {IEEE} {IACC}.
  Bangalore, India, pp. 56--60.

\bibitem[{Manivannan et~al.(2017)Manivannan, Cobb, Burgess, and
  Trucco}]{manivannan_sub-category_2017}
Manivannan, S., Cobb, C., Burgess, S., Trucco, E., 2017. Sub-category
  classifiers for multiple-instance learning and its application to retinal
  nerve fiber layer visibility classification. IEEE Trans Med Imaging (in
  press).

\bibitem[{Melendez et~al.(2015)Melendez, van Ginneken, Maduskar, Philipsen,
  Reither, Breuninger, Adetifa, Maane, Ayles, and
  S{\'a}nchez}]{melendez_novel_2015}
Melendez, J., van Ginneken, B., Maduskar, P., Philipsen, R. H. H.~M., Reither,
  K., Breuninger, M., Adetifa, I. M.~O., Maane, R., Ayles, H., S{\'a}nchez,
  C.~I., Jan. 2015. A novel multiple-instance learning-based approach to
  computer-aided detection of tuberculosis on chest {X}-rays. IEEE Trans Med
  Imaging 34~(1), 179--192.

\bibitem[{Nesterov(1983)}]{nesterov_method_1983}
Nesterov, Y., 1983. A method of solving a convex programming problem with
  convergence rate {O}(1/sqr(k)). Soviet Math Doklady 27, 372--376.

\bibitem[{Nielsen(2015)}]{nielsen_how_2015}
Nielsen, M.~A., 2015. How the backpropagation algorithm works. In: Neural
  networks and deep learning. Determination Press, Ch.~2.

\bibitem[{Prentasic and Loncaric(2015)}]{prentasic_detection_2015}
Prentasic, P., Loncaric, S., Sep. 2015. Detection of exudates in fundus
  photographs using convolutional neural networks. In: Proc {ISPA}. Zabreb,
  Croatia, pp. 188--192.

\bibitem[{Quellec et~al.(2016{\natexlab{a}})Quellec, Bazin, Cazuguel, Delafoy,
  Cochener, and Lamard}]{quellec_suitability_2016}
Quellec, G., Bazin, L., Cazuguel, G., Delafoy, I., Cochener, B., Lamard, M.,
  Apr. 2016{\natexlab{a}}. Suitability of a low-cost, handheld, nonmydriatic
  retinograph for diabetic retinopathy diagnosis. Transl Vis Sci Technol 5~(2),
  16.

\bibitem[{Quellec et~al.(2017)Quellec, Cazuguel, Cochener, and
  Lamard}]{quellec_multiple-instance_2017}
Quellec, G., Cazuguel, G., Cochener, B., Lamard, M., 2017. Multiple-instance
  learning for medical image and video analysis. IEEE Rev Biomed Eng (in
  press).

\bibitem[{Quellec et~al.(2016{\natexlab{b}})Quellec, Lamard, Cozic, Coatrieux,
  and Cazuguel}]{quellec_multiple-instance_2016}
Quellec, G., Lamard, M., Cozic, M., Coatrieux, G., Cazuguel, G., Jul.
  2016{\natexlab{b}}. Multiple-instance learning for anomaly detection in
  digital mammography. IEEE Trans Med Imaging 35~(7), 1604--1614.

\bibitem[{Quellec et~al.(2016{\natexlab{c}})Quellec, Lamard, Erginay, Chabouis,
  Massin, Cochener, and Cazuguel}]{quellec_automatic_2016}
Quellec, G., Lamard, M., Erginay, A., Chabouis, A., Massin, P., Cochener, B.,
  Cazuguel, G., Apr. 2016{\natexlab{c}}. Automatic detection of referral
  patients due to retinal pathologies through data mining. Med Image Anal 29,
  47--64.

\bibitem[{Russakovsky et~al.(2015)Russakovsky, Deng, Su, Krause, Satheesh, Ma,
  Huang, Karpathy, Khosla, Bernstein, Berg, and
  Fei-Fei}]{russakovsky_imagenet_2015}
Russakovsky, O., Deng, J., Su, H., Krause, J., Satheesh, S., Ma, S., Huang, Z.,
  Karpathy, A., Khosla, A., Bernstein, M., Berg, A.~C., Fei-Fei, L., Apr. 2015.
  {ImageNet} large scale visual recognition challenge. Int J Comput Vis
  115~(3), 211--252.

\bibitem[{Samek et~al.(2016)Samek, Binder, Montavon, Lapuschkin, and
  M{\"u}ller}]{samek_evaluating_2016}
Samek, W., Binder, A., Montavon, G., Lapuschkin, S., M{\"u}ller, K.~R., 2016.
  Evaluating the visualization of what a deep neural network has learned. IEEE
  Trans Neural Netw Learn Syst (in press).

\bibitem[{Simonyan et~al.(2014)Simonyan, Vedaldi, and
  Zisserman}]{simonyan_deep_2014}
Simonyan, K., Vedaldi, A., Zisserman, A., Apr. 2014. Deep inside convolutional
  networks: visualising image classification models and saliency maps. In:
  {ICLR} {Workshop}. Calgary, Canada.

\bibitem[{Srivastava et~al.(2015)Srivastava, Cheng, Wong, and
  Liu}]{srivastava_using_2015}
Srivastava, R., Cheng, J., Wong, D., Liu, J., Apr. 2015. Using deep learning
  for robustness to parapapillary atrophy in optic disc segmentation. In: Proc
  {ISBI}. New York, NY, USA, pp. 768--771.

\bibitem[{Tibshirani(1996)}]{tibshirani_regression_1996}
Tibshirani, R., 1996. Regression shrinkage and selection via the lasso. J Royal
  Statist Soc B 58~(1), 267--288.

\bibitem[{van Grinsven et~al.(2016)van Grinsven, van Ginneken, Hoyng, Theelen,
  and Sanchez}]{van_grinsven_fast_2016}
van Grinsven, M., van Ginneken, B., Hoyng, C., Theelen, T., Sanchez, C., Feb.
  2016. Fast convolutional neural network training using selective data
  sampling: {Application} to hemorrhage detection in color fundus images. IEEE
  Trans Med Imaging.

\bibitem[{Wilkinson et~al.(2003)Wilkinson, Ferris, Klein, Lee, Agardh, Davis,
  Dills, Kampik, Pararajasegaram, and Verdaguer}]{wilkinson_proposed_2003}
Wilkinson, C.~P., Ferris, F.~L., Klein, R.~E., Lee, P.~P., Agardh, C.~D.,
  Davis, M., Dills, D., Kampik, A., Pararajasegaram, R., Verdaguer, J.~T., Sep.
  2003. Proposed international clinical diabetic retinopathy and diabetic
  macular edema disease severity scales. Ophthalmology 110~(9), 1677--1682.

\bibitem[{Winder et~al.(2009)Winder, Morrow, McRitchie, Bailie, and
  Hart}]{winder_algorithms_2009}
Winder, R.~J., Morrow, P.~J., McRitchie, I.~N., Bailie, J.~R., Hart, P.~M.,
  Dec. 2009. Algorithms for digital image processing in diabetic retinopathy.
  Comput Med Imaging Graph 33~(8), 608--622.

\bibitem[{Worrall et~al.(2016)Worrall, Wilson, and
  Brostow}]{worrall_automated_2016}
Worrall, D.~E., Wilson, C.~M., Brostow, G.~J., Oct. 2016. Automated retinopathy
  of prematurity case detection with convolutional neural networks. In: Proc
  {MICCAI} {Deep} {Learning} {Workshop}. Lecture {Notes} in {Computer}
  {Science}. pp. 68--76.

\bibitem[{Yang et~al.(2013)Yang, Lu, Fang, and Yang}]{yang_effective_2013}
Yang, N., Lu, H.-C., Fang, G.-L., Yang, G., 2013. An effective framework for
  automatic segmentation of hard exudates in fundus images. J Circuit Syst Comp
  22~(1).

\bibitem[{Yosinski et~al.(2015)Yosinski, Clune, Nguyen, Fuchs, and
  Lipson}]{yosinski_understanding_2015}
Yosinski, J., Clune, J., Nguyen, A., Fuchs, T., Lipson, H., Jul. 2015.
  Understanding neural networks through deep visualization. In: Proc {ICML}
  {DL} {Works}. Lille, France.

\bibitem[{Zeiler and Fergus(2014)}]{zeiler_visualizing_2014}
Zeiler, M.~D., Fergus, R., Sep. 2014. Visualizing and understanding
  convolutional networks. In: Proc {ECCV}. Zurich, Switzerland, pp. 818--833.

\end{thebibliography}

\appendix

\section{Popular ConvNet Operators}
\label{app:PopularConvNetOperators}

This appendix describes the operators used by the ConvNets evaluated in this paper. Their backward first-order derivatives and their forward second-order derivatives are given in \ref{app:ForwardSecondOrderDerivativesPopularConvNetOperators}.

\subsection{Convolutional Layer \textit{(Conv)}}

Let $w_l \times h_l$ denote the size of the sliding window and let $s_l$ denote its stride, i.e. the offset between two sliding window locations. The value of $D^{(l)}$ in the $c$\textsuperscript{th} activation map of the $n$\textsuperscript{th} image, at the $(x, y)$ coordinate, is obtained by a cross-correlation product between $D^{(l-1)}$, inside a sliding window centered on $(s_l x, s_l y)$, and a tensor $\Omega^{(l)}$ of $C_l$ filters with dimensions $w_l \times h_l \times C_{l-1}$ each. A bias $b^{(l)}_{x, y, c}$ is added to the product and a nonlinear activation function $a_l$ is applied to the sum:
\begin{equation}
  \label{eq:conv}
  D^{(l)}_{n, x, y, c} = a_l \left(\sum_{u = -\tfrac{w_l}{2}}^{w_l-\tfrac{w_l}{2}}\sum_{v = -\tfrac{h_l}{2}}^{h_l-\tfrac{h_l}{2}}\sum_{d = 1}^{C_{l-1}}{\Omega^{(l)}_{u, v, d, c}} D^{(l-1)}_{n, s_l x + u, s_l y + v, d} + b^{(l)}_{x, y, c} \right)  \;.
\end{equation}
Biases are generally tied, meaning that $b^{(l)}_{x, y, c} = b^{(l)}_{c}$, otherwise they are said to be untied.

\subsection{Activation Functions}

Currently, the most popular activation functions are rectifiers and leaky rectifiers \citep{he_delving_2015}, which are much less computationally intensive than the traditional sigmoid function for instance. They can be expressed as follows:
\begin{equation}
  \label{eq:rectifier}
  r_\alpha(x) = \max(\alpha x, x), \forall x \in \mathbb{R},
\end{equation}
where $\alpha \in[0, 1[$ is usually small: $\alpha = 0.01$, typically, or $\alpha = 0$ in the original rectifier function. Leaky rectifiers with large $\alpha$ values (e.g. $\alpha = 0.33$) are called `very leaky rectifiers'.

\subsection{Dense or Fully-Connected Layers \textit{(Dense)}}

Dense layers are a special case of convolutional layers where $w_l$ = $W_{l-1}$ and $h_l$ = $H_{l-1}$, so the dimensions of $D^{(l)}$ are $N \times 1 \times 1 \times C_l$, where $C_l$ is the number of neurons / filters in layer $l$.

\subsection{Pooling Layer \textit{(MaxPool, MeanPool or RMSPool)}}

Pooling layers also rely on a $w_l \times h_l$ sliding window with a stride of $s_l$. Those layers replace the content of the sliding window in the input data tensor by a single value in the output tensor. With a stride greater than 1, a down-sampling operation is performed. Typical pooling operators are the maximum, the mean and the root mean square (RMS). Let $\mathcal{N}_l(x,y)=\{(s_l x + u, s_l y + v) \text{ s.t. } u = -\tfrac{w_l}{2}, ..., w_l-\tfrac{w_l}{2}, v = -\tfrac{h_l}{2}, ..., h_l-\tfrac{h_l}{2}\}$ denote the neighborhood of pixel location $(x, y)$ inside $D^{(l-1)}$. A MaxPool layer computes:
\begin{equation}
  \label{eq:maxPool}
  D^{(l)}_{n, x, y, c} = \max_{(u,v) \in \mathcal{N}_l(x,y)}{D^{(l-1)}_{n, u, v, c}}
\end{equation}
and a MeanPool layer computes:
\begin{equation}
  \label{eq:meanPool}
  D^{(l)}_{n, x, y, c} = \frac{1}{w_l \times h_l}\sum_{(u,v) \in \mathcal{N}_l(x,y)}{D^{(l-1)}_{n, u, v, c}} \;\;.
\end{equation}
Note that the number of activation maps is unchanged by those operators: $C_l = C_{l-1}$. Finally, an RMSPool layer simply derives from a MeanPool layer as follows:
\begin{equation}
  RMSPool\left( D^{(l)} \right) = \sqrt{MeanPool\left( { D^{(l)} }^2 \right)} \;\;.
\end{equation}

\subsection{Dropout and Maxout Layers}

Dropout is a popular regularization technique \citep{hinton_improving_2012}. During each training iteration, a random selection of filters from layer $l+1$ (one filter in $p$) are `dropped': their input and output connections are temporarily removed. The goal is to train multiple models, where each model is a `thinned' version of the ConvNet. For improved performance, a Maxout layer can be placed in position $l+2$: this operator simply returns the maximum output among subsets of $p$ filters from layer $l+1$ \citep{goodfellow_maxout_2013}.

\section{Forward Second-Order Derivatives of Popular ConvNet Operators}
\label{app:ForwardSecondOrderDerivativesPopularConvNetOperators}

\subsection{Forward Second-Order Derivatives for Cross-Correlation}

We first discuss the main building-block of ConvNets, namely cross-correlation between data $D^{(l-1)}$ and filter weights $\Omega^{(l)}$, used by convolutional and dense layers [see Equation (\ref{eq:conv})]. Bias addition, which is trivial, is not discussed and activation is discussed in the following section. Because cross-correlation has two inputs ($D^{(l-1)}$ and $\Omega^{(l)}$), two backward first-order derivative functions need to be computed:
\begin{equation}
  \label{eq:convGradInput}
  \left( \frac{\partial \mathcal L_L}{\partial D^{(l-1)}} \right)_{n,x,y,d} = \displaystyle\sum_{u = -\tfrac{w_l}{2}}^{w_l-\tfrac{w_l}{2}}\sum_{v = -\tfrac{h_l}{2}}^{h_l-\tfrac{h_l}{2}}\sum_{c = 1}^{C_l}{ \Omega^{(l)}_{x,y,d,c} \left( \frac{\partial \mathcal L_L}{\partial D^{(l)}} \right)_{n, s_l x - u, s_l y - v, c} }  \;\;,
\end{equation}
\begin{equation}
  \label{eq:convGradFilter}
  \left( \frac{\partial \mathcal L_L}{\partial \Omega^{(l)}} \right)_{x,y,d,c} = \displaystyle\sum_{u = -\tfrac{w_l}{2}}^{w_l-\tfrac{w_l}{2}}\sum_{v = -\tfrac{h_l}{2}}^{h_l-\tfrac{h_l}{2}}\sum_{n = 1}^N{ \left( \frac{\partial \mathcal L_L}{\partial D^{(l)}} \right)_{n,x,y,c} D^{(l-1)}_{n,s_lx+u,s_ly+v,d} } \;\;.
\end{equation}
These equations derive from the chain rule of derivation [see Equation (\ref{eq:chainRule2})] and the differentiation of a cross-correlation product  \citep{nielsen_how_2015}. We can see that the cross-correlation between $\Omega^{(l)}$ and $D^{(l-1)}$ in the forward transform [see Equation (\ref{eq:conv})] becomes an actual convolution product between $\Omega^{(l)}$ and $\frac{\partial \mathcal L_L}{\partial D^{(l)}}$ in the backward first-order derivative function [see Equation (\ref{eq:convGradInput})]. As expected, one can verify that it becomes a cross-correlation product again, between $\Omega^{(l)}$ and $\frac{\partial \mathcal L_0}{\partial D^{(l-1)}}$, in the forward second-order derivative function:
\begin{equation}
  \label{eq:convGradGradInput}
  \left( \frac{\partial \mathcal L_0}{\partial D^{(l)}} \right)_{n, x, y, c} = \displaystyle\sum_{u = -\tfrac{w_l}{2}}^{w_l-\tfrac{w_l}{2}}\sum_{v = -\tfrac{h_l}{2}}^{h_l-\tfrac{h_l}{2}}\sum_{d = 1}^{C_{l-1}}{\Omega^{(l)}_{u, v, d, c}} \left( \frac{\partial \mathcal L_0}{\partial D^{(l-1)}} \right)_{n, s_l x + u, s_l y + v, d}  \;\;.
\end{equation}
The gradient of the loss function with respect to $\Omega^{(l)}$ is a cross-correlation product between $\tfrac{\partial \mathcal L_L}{\partial D^{(l)}}$, playing the role of a filter, and the input data [see Equation (\ref{eq:convGradFilter})]. One can verify that the same applies during the forward pass, where $\frac{\partial \mathcal L_0}{\partial D^{(l-1)}}$ plays the role of the input data:
\begin{equation}
  \label{eq:convGradGradFilter}
  \left( \frac{\partial \mathcal L_0}{\partial \Omega^{(l)}} \right)_{x,y,d,c} = \displaystyle\sum_{u = -\tfrac{w_l}{2}}^{w_l-\tfrac{w_l}{2}}\sum_{v = -\tfrac{h_l}{2}}^{h_l-\tfrac{h_l}{2}}\sum_{n = 1}^N{ \left( \frac{\partial \mathcal L_L}{\partial D^{(l)}} \right)_{n,x,y,c} \left( \frac{\partial \mathcal L_0}{\partial D^{(l-1)}} \right)_{n,s_lx+u,s_ly+v,d} }.
\end{equation}

\subsection{Forward Second-Order Derivatives for the Leaky Rectifier}

Let $y = r_\alpha(x) = \max(\alpha x, x)$ with $\alpha \in [0, 1[$ [see Equation (\ref{eq:rectifier})]. Because $r_\alpha$ simply is a piecewise linear operator, the backward first-order derivatives for $r_\alpha$ are given by:
\begin{equation}
  \label{eq:rectifierGrad}
  \frac{\partial \mathcal L_L}{\partial x} = \left\lbrace
  \begin{array}{ll}
    0 & \text{if } \alpha = 0 \text{ and }  x < 0 \\
    \frac{1}{\alpha} \frac{\partial \mathcal L_L}{\partial y} & \text{if } \alpha > 0 \text{ and }  x < 0 \\
    \frac{\partial \mathcal L_L}{\partial y} & \text{if } x \geq 0
  \end{array}
  \right.
\end{equation}
and the forward second-order derivatives are given by:
\begin{equation}
  \label{eq:rectifierGradGrad}
  \frac{\partial \mathcal L_0}{\partial y} = \left\lbrace
  \begin{array}{ll}
    \alpha \frac{\partial \mathcal L_0}{\partial x} & \text{ if } x < 0 \\
    \frac{\partial \mathcal L_0}{\partial x} & \text{ if } x \geq 0
  \end{array}
  \right. .
\end{equation}
Note that the test for choosing the multiplicative factor (1 or $\alpha$) is always triggered by the operator's input ($x$). The same applies to the MaxPool operator below.

\subsection{Forward Second-Order Derivatives for MaxPool}

The backward first-order derivatives for MaxPool [see Equation (\ref{eq:maxPool})] are given by:
\begin{equation}
  \label{eq:maxPoolGrad}
  \left(\frac{\partial \mathcal L_L}{\partial D^{(l-1)}}\right)_{n,x,y,c} = \sum_{\substack{u, v\\ \text{s.t. } D^{(l)}_{n,u,v,c} = D^{(l-1)}_{n,x,y,c}}}{ \left(\frac{\partial \mathcal L_L}{\partial D^{(l)}}\right)_{n,u,v,d} }  \;\;.
\end{equation}
This means that the errors are backpropagated to the winning neuron inside each sliding window location. One can verify that the forward second-order derivatives are given by:
\begin{equation}
  \label{eq:maxPoolGradGrad}
  \left(\frac{\partial \mathcal L_0}{\partial D^{(l)}}\right)_{n,x,y,c} = \left(\frac{\partial \mathcal L_0}{\partial D^{(l-1)}}\right)_{\argmax\left( D^{(l)}_{n,x,y,c} \right)}  \;\;,
\end{equation}
where $\argmax\left( D^{(l)}_{n,x,y,c} \right)$ returns the index of the winning neuron inside the corresponding sliding window in $D^{(l-1)}$.

\subsection{Forward Second-Order Derivatives for MeanPool}

The backward first-order derivatives for MeanPool [see Equation (\ref{eq:meanPool})] are given by:
\begin{equation}
  \label{eq:meanPoolGrad}
  \left(\frac{\partial \mathcal L_L}{\partial D^{(l-1)}}\right)_{n,x,y,c} = \frac{1}{w_l \times h_l} \sum_{\substack{u, v\\ \text{s.t. } (x,y) \in \mathcal{N}_l(u,v)}}{ \left(\frac{\partial \mathcal L_L}{\partial D^{(l)}}\right)_{n,u,v,c} } \;\;.
\end{equation}
This means that, during backpropagation, the errors are equally distributed to all neurons inside each sliding window location. The forward second-order derivatives are a special case of Equation (\ref{eq:convGradGradInput}), where $W_l$ is a mean filter; it is given by:
\begin{equation}
  \label{eq:meanPoolGradGrad}
  \left(\frac{\partial \mathcal L_0}{\partial D^{(l)}}\right)_{n,x,y,c} = \frac{1}{w_l \times h_l}\sum_{(u,v) \in \mathcal{N}_l(x,y)}{\left(\frac{\partial \mathcal L_0}{\partial D^{(l-1)}}\right)_{n, u, v, c}}  \;\;.
\end{equation}
In other words, the forward second-order derivative function for MeanPool is MeanPool itself.

\subsection{Forward Second-Order Derivatives for Dropout and Maxout}

Dropout does not need to be addressed specifically as it simply alters the network temporarily: the above first-order and second-order derivatives are simply computed in the thinned network. As for Maxout, it is addressed similarly to the other maximum-based operators (leaky rectifiers and MaxPool).

\end{document}